%% file: acl_latex.tex
\pdfoutput=1
\documentclass[11pt]{article}

\usepackage[]{acl}
\usepackage{graphicx}
\usepackage{times}
\usepackage{latexsym}

\usepackage[T1]{fontenc}
\usepackage[utf8]{inputenc}

\usepackage{microtype}
\usepackage{inconsolata}
\usepackage{booktabs} % !
\usepackage{multirow}
\usepackage{amsmath}
\usepackage{amsthm}
\usepackage{amssymb}
\usepackage{amsfonts}
\usepackage{pifont}
\usepackage{array}
\usepackage{xcolor}
\usepackage{hyperref} %url换行
\usepackage{subcaption} % \linewidth
\usepackage{graphicx}
\usepackage{listings}
\usepackage{algorithm}
\usepackage{algorithmicx}
\usepackage{algpseudocode}
\usepackage{lipsum} % for dummy text
\usepackage[most]{tcolorbox}
\usepackage[normalem]{ulem}
\newcommand{\bblack}[1]{{\textbf{\color{black}{#1}}}}
\newcommand{\okmark}{{\textbf{\textcolor[rgb]{0.1, 0.5, 0.1}{$\checkmark$}}}}
\newcommand{\ngmark}{{\textbf{\color{red}{\ding{55}}}}}
\definecolor{nmgray}{RGB}{229,229,229}
\definecolor{underlinegray}{RGB}{197,197,197}
\definecolor{introblue}{RGB}{0,176,240}
\definecolor{introgreen}{RGB}{0,203,134}
\definecolor{introgreen2}{RGB}{139,243,206}
\newtcolorbox{mybox}[1][]{
    width=\columnwidth,
    colback=nmgray!75!white, 
    colframe=nmgray!75!white, 
    coltitle=black,       % change this line, color of the title text
    fonttitle=\bfseries,  % change this line, style of the title text
    boxsep=2pt,
    left=4pt,
    right=2pt,
    top=2pt,
    bottom=2pt,
    fontupper=\linespread{0.9}\selectfont,
    title=#1,
    before upper=\ttfamily,
}

\title{Mitigating Boundary Ambiguity and Inherent Bias for \\ Text Classification in the Era of Large Language Models}

\author{
    Zhenyi Lu\textsuperscript{1,2}\thanks{\quad Equal contribution.}, Jie Tian\textsuperscript{1,2}\footnotemark[1], Wei Wei\textsuperscript{1,2}\thanks{\quad Corresponding authors: Wei Wei and Dangyang Chen.}, Xiaoye Qu\textsuperscript{1,3}, \\
    \textbf{Yu Cheng}\textsuperscript{4}, \textbf{Wenfeng Xie}\textsuperscript{5}, \textbf{Dangyang Chen}\textsuperscript{5}\footnotemark[2] \\
    \textsuperscript{1} Cognitive Computing and Intelligent Information Processing (CCIIP) Laboratory, \\
    School of Computer Science and Technology, Huazhong University of Science and Technology, \\
    \textsuperscript{2} Joint Laboratory of HUST and Pingan Property \& Casualty Research (HPL), \\
    \textsuperscript{3} Shanghai AI Laboratory, \textsuperscript{4} The Chinese University of Hong Kong, \\
    \textsuperscript{5} Ping An Property \& Casualty Insurance Company of China, Ltd. \\
    \texttt{ \{luzhenyi529,jaytian.china\}@gmail.com, \{weiw, xiaoye\}@hust.edu.cn} \\
    \texttt{chengyu@cse.cuhk.edu.hk, julian\_wind@163.com, chendangyang273@pingan.com.cn}
}

\begin{document}

\maketitle

\begin{abstract}
% Even with extended long context, these models struggle with fuzzy decision boundaries and subtle label distinctions. 
% this paper aims to shed light on their limitations in handling such tasks through extensive empirical analysis.
% It is observed that LLMs tend to exhibit a decline in performance as the number of classification options increases.
% Classifying texts with numerous potential classes (> 50 options) is a crucial yet frequently encountered task in real-life scenarios. 
% Despite their impressive language comprehension capabilities, 
% LLMs still encounter challenges in distinguishing between a growing number of options and subtle differences between them.
% Furthermore, they favor specific optional positions or selectively retain or discard tokens.  
% ------------------------------------------------------------------------------------------------

Text classification is a crucial task encountered frequently in practical scenarios, yet it is still under-explored in the era of large language models (LLMs). 
% This study reveals that LLMs are vulnerable to changes in the number and arrangement of options in text classification, exhibiting the fuzzy decision boundaries and inherent biases towards specific tokens and positions.
This study shows that LLMs are vulnerable to changes in the number and arrangement of options in text classification. Our extensive empirical analyses reveal that the key bottleneck arises from ambiguous decision boundaries and inherent biases towards specific tokens and positions.
To mitigate these issues, we make the first attempt and propose a novel two-stage classification framework for LLMs. 
Our approach is grounded in the empirical observation that pairwise comparisons can effectively alleviate boundary ambiguity and inherent bias. 
% thereby harnessing the classification capabilities of LLMs. the subsequent comparison process.
Specifically, we begin with a self-reduction technique to efficiently narrow down numerous options, which contributes to reduced decision space and a faster comparison process. 
Subsequently, pairwise contrastive comparisons are employed in a chain-of-thought manner to draw out nuances and distinguish confusable options, thus refining the ambiguous decision boundary.
Extensive experiments on four datasets (Banking77, HWU64, LIU54, and Clinic150) verify the effectiveness of our framework. Furthermore, benefitting from our framework, various LLMs can achieve consistent improvements. Our code and data are available in \url{https://github.com/Chuge0335/PC-CoT}.

\end{abstract}

\section{Introduction}

% Chain-of-Thought (CoT) has demonstrated notable success in enhancing the performance of large language models (LLMs).
% However, its applicability in identifying intricate labels and nuanced reasoning within real-world text classification problems remains relatively unexplored. 
% This arises from the inherent complexities posed by practical scenarios, including an abundance of label choices, custom-interpreted labels, and subtle distinctions between labels. 
% In such contexts, the straightforward application of LLMs for classification proves non-trivial.
% the observation that pairwise comparison can unlock the potential of LLMs for classification and migrate among options .
% multi-class
% limitation: hierarchical classification, multi-label classification, extreme classification (thousands of classes such that exceed the context length of LLMs), clustering future work to explored
%  which may be an inherent behavioral bias of LLMs that cannot be addressed by simple prompt engineering.

\begin{figure}[t]
    \centering
    \includegraphics[width=1\columnwidth]{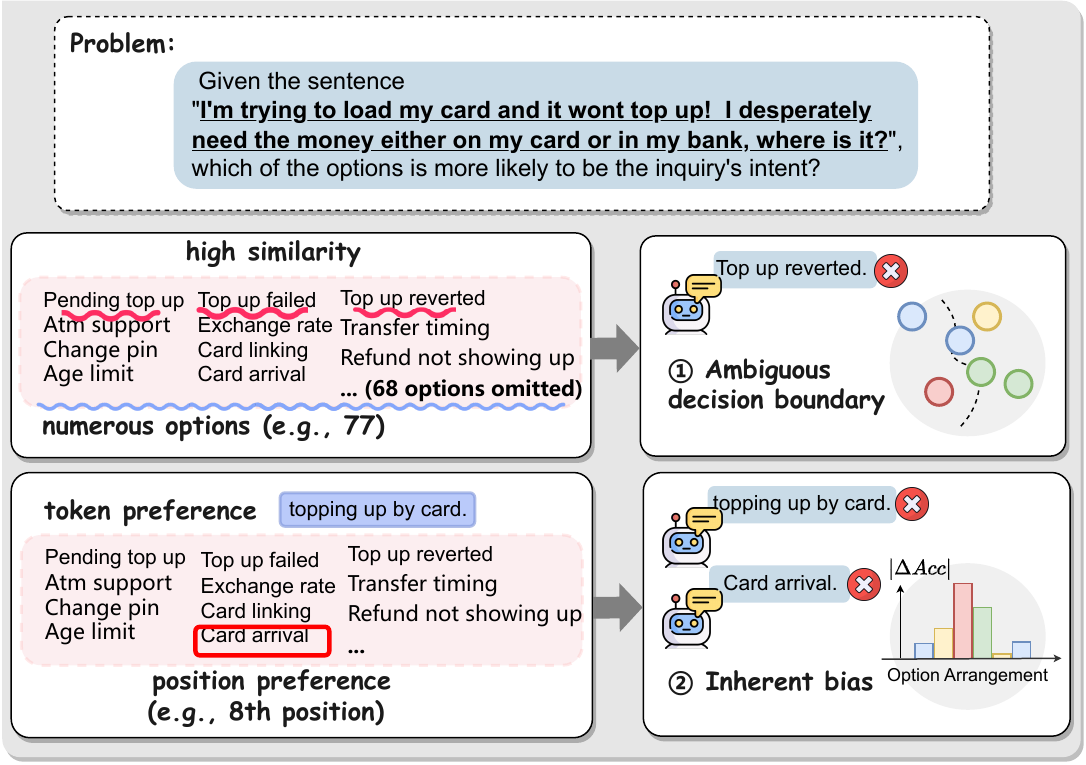}
    \caption{
        LLMs in text classification suffer from two main challenges: (a) ambiguous decision boundary, which arises from the explosion of options and their similarity. (b) inherent bias, favoring specific optional positions or tokens. These challenges 
        % make LLMs vulnerable to changes in the number and arrangement of options in text classification tasks, 
        lead to decreased classification accuracy.}
    \label{fig:sample}
    \vspace{-5mm}
\end{figure}

% These approaches effectively enhance the reasoning abilities of LLMs by introducing step-by-step reasoning processes in immediate contexts.
% However, the specific Chain-of-Thought (CoT) reasoning associated with complex text classification, such as identifying underly intent from hundreds of labels, remains a less-explored area. 
% The complex text classification problem has its own distinct challenge need to handle: \textbf{options explosion} and \textbf{boundary ambiguity}, 

%  class imbalance problem
Text classification, as one of the most fundamental and common tasks in natural language processing (NLP), 
has a wide spectrum of applications~\citep{qu2019adversarial,2023TARTImprovedFewshot,2023ContrastNetContrastiveLearning,zaremba2023chatgpt,labonne2023spamt5}.
A common approach for text classification involves supervised fine-tuning on a specific dataset, which can be costly and may suffer from generalization issues~\citep{2023PretrainingIntentAwareEncoders,2023BYOCPersonalizedFewShot,2023EffectiveStructuredPrompting}. 
Recent Large Language Models (LLMs), such as ChatGPT~\citep{Chatgpt}, LLaMA2~\citep{touvron2023llama2} and Qwen~\citep{bai2023qwen}, 
have consistently exhibited remarkable language comprehension capabilities and performance emergence~\cite {wei2022emergent}.
Such advancement has introduced a new paradigm of classification without the need for finetuning. 
LLMs can now make selections in a zero-shot manner, or employ techniques like few-shot demonstrations~\citep{brown2020language} and chain-of-thought (CoT)~\citep{kojima2023large}.

Unfortunately, we observe that LLMs exhibit vulnerability to changes in the number and arrangement of options in text classification tasks. {Our extensive empirical evaluation, involving 11 LLMs ranging from 1B to 70B parameters and classification tasks with over 50 options}, demonstrates that \textbf{boundary ambiguity} and \textbf{inherent bias} are prevalent issues in various large language models. \textit{These issues cannot be effectively mitigated by long-context capabilities, simple prompting strategies, or single increasing the model size.}
% ------------------------------------------------
LLMs struggle with distinguishing among a growing number of options, a phenomenon called \textit{boundary ambiguity}. For example, gpt-3.5-turbo achieves a zero-shot accuracy of only $32.51$\% when choosing the correct option from $60$ categories, compared to $94.29$\% for two categories. LLMs also exhibit \textit{inherent bias}, favoring specific positions or selectively retaining or discarding tokens in classification tasks. Placing the correct option consistently in the first position results in performance fluctuations of $35.29$\% for LLaMA2-70B-Chat and $38.52$\% for Qwen-72B-Chat. Furthermore, LLaMA2 shows an abnormal bias towards selecting ``movies'' and ``convert'' in the LIU54 benchmark.

% Previous work~\citep{2023ReasoningImplicitSentiment,sun-etal-2023-text} has primarily focused on simpler classification problems with only 2-4 options, making them impractical for addressing such ambiguities and biases.

To address these challenges for LLMs, our intuition is that LLMs often have the highest accuracy in \textbf{pairwise} classification, and focusing on two options at a time can significantly reduce the number of tokens and positions considered, potentially alleviating bias issues. 
This divide-and-conquer approach also mirrors human decision-making with multiple options, where we first eliminate less likely options and then carefully differentiate between similar ones. The latter process often involves juxtaposing confusable objects pairwisely to assess their similarities and differences.

% which initially narrow down numerous options and further refine the selection through careful comparison in a pairwise manner.
In view of this, we design a general two-stage LLM classification framework consisting of reduction and comparison stages.
Within such a framework, we introduce two novel self-reduction techniques for the reduction process: one iteratively isolates the most probable option from candidates, and the other induces distinct option ``windows'' via clustering, operating within these windows to avoid similarity confusion, termed as ITR and CBWR. 
Additionally, we propose PC-CoT, an innovative contrastive chain-of-thought technique, to draw out nuances and distinguish features among confusable options, refining the ambiguous decision boundary. {This type of detailed comparison} encourages LLMs to conduct more in-depth analyses of actual content, rather than making shortcuts based solely on position or tokens.
We conduct extensive experiments to demonstrate that our framework is consistently effective for various LLMs to combat ambiguity and bias, even when the number of options increases to $150$. Notably, {our method achieves a relative improvement of 54.1\% in average accuracy compared to the full-option zero-shot performance of gpt-3.5-turbo}.
It significantly reduces the token bias of LLaMA-70B-Chat, improving the relative performance by 36.88\%.

To sum up, our contributions are as follows: 
(1) Our work provides extensive empirical analyses (covering 11 LLMs) to explore the vulnerability of LLMs in option number and arrangement in text classification. We identify that their ambiguous decision boundaries and inherent biases on tokens and positions remain a main challenge. 
(2) To address these issues, we propose utilizing pairwise comparison to combat such ambiguity and bias and develop an innovative two-stage classification framework for LLMs. 
Within the framework, novel self-reduction algorithms and a contrastive elimination strategy are introduced. (3) We carry out comprehensive experiments on four primary benchmarks along with diverse LLMs, demonstrating the effectiveness of our framework.

\begin{figure}[h]
    \centering
    \includegraphics[width=\columnwidth]{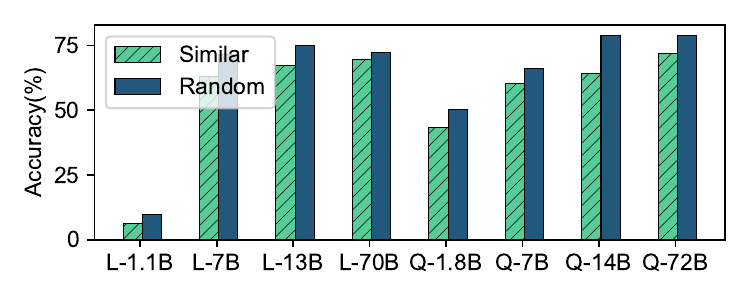}
    \caption{The classification accuracy decreases across various LLMs as the similarity among options increases. ``L'' denotes the LLaMA-2-Chat series, while ``Q'' represents the Qwen-Chat series.}
    \label{fig:sim}
\end{figure}

\begin{figure*}[t]

% medskip周围不能加换行
\begin{minipage}{0.49\linewidth}
    \centering
    \begin{subfigure}{1\linewidth}
        \includegraphics[width=1\linewidth]{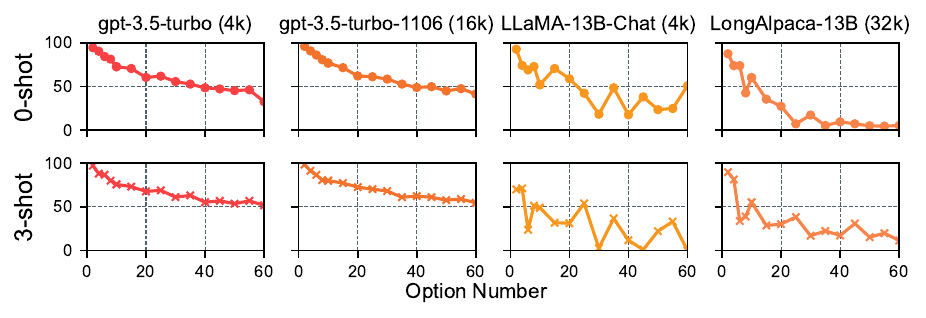}
        \caption{LLMs classification accuracy in relation to the number of options consistently demonstrates a decline.}
        \label{fig:a}
    \end{subfigure}
    \begin{subfigure}{1\linewidth}
        \includegraphics[width=1\linewidth]{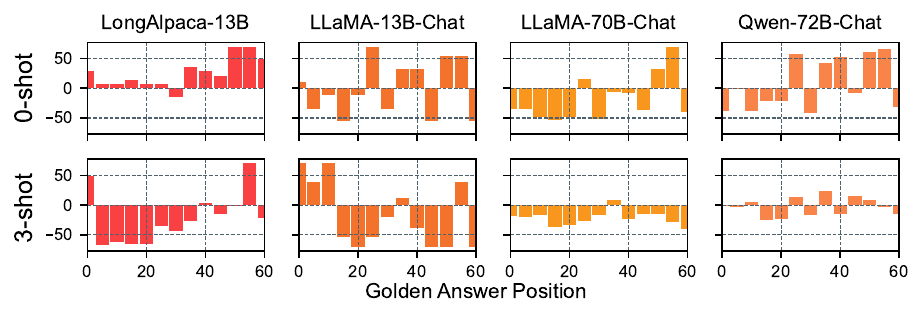}
        \caption{
        % The change rate of classification accuracy after moving the correct option to a fixed position compared to the original accuracy. 
        The relative performance change rate for fixed correct options in a position (range from 0 to 60) compared to random configuration performance.
        Simple option arrangement leads to dramatic performance fluctuations in both zero-shot and few-shot scenarios.} 
        % after moving the correct option to a fixed position
        \label{fig:b}
    \end{subfigure}
\end{minipage} %\par
\hspace{5mm}
\begin{minipage}{0.49\linewidth}
    \begin{subfigure}{1\linewidth}
        \centering
        \includegraphics[width=0.8\linewidth]{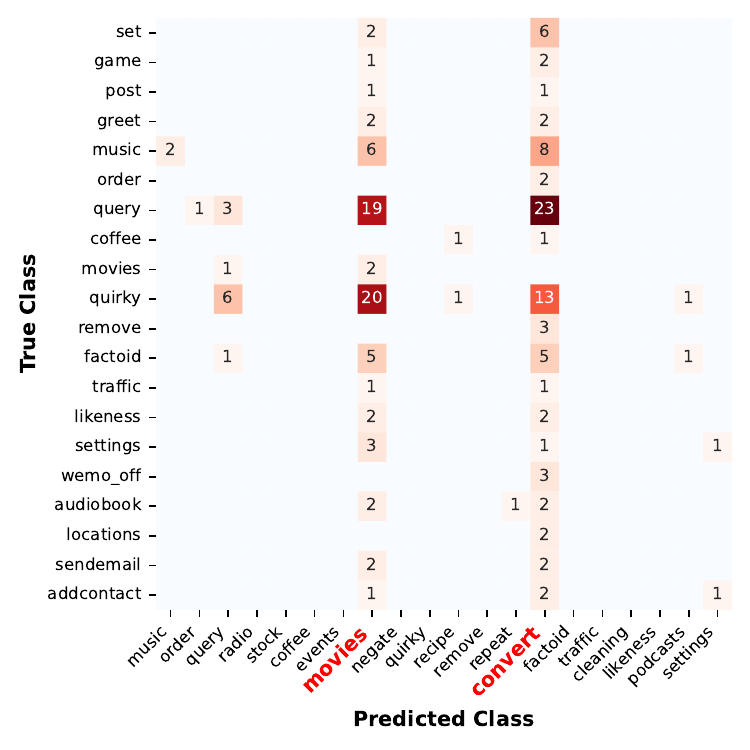}
        \caption{The confusion matrix sheds light on the token bias of LLaMA-2 in the LIU54 dataset. Notably, the examination reveals a pronounced bias of LLaMA-2-70B-Chat towards the tokens ``movies'' and ``convert''. }
        \label{fig:c}
    \end{subfigure}
\end{minipage}
\caption{The influence of option number and arrangement on LLM performance. Full results are in the Appendix~\ref{sec:0}. }
    % \vspace{-3mm}
\label{fig:3body}
\end{figure*}

\section{Preliminary Study}
\label{sec.2}

\paragraph{Setup}

% We conduct an extensive evaluation across various LLMs to establish a preliminary comprehension of factors influencing classification accuracy. 
% We experiment with popular LLM families including gpt-3.5 series, as well as open-source models like the LLaMA2-Chat series and the Qwen-Chat series.
% Our evaluation is conducted on the Banking77 dataset and LIU54 dataset (details in  Section~\ref{sec.4}).
% To reduce experimental costs, we sample only the challenging portions of these datasets. 
% Due to limited space, we will summarize the main results and findings in the main text, and move the dataset sample detail and more results in the Appendix~\ref{sec:0}.

We evaluate various LLMs to understand factors influencing classification accuracy. Our experiments include the gpt-3.5 series, LLaMA2-Chat, and Qwen-Chat models. We use the challenging portions of the Banking77 and LIU54 datasets (details in Section~\ref{sec.4}) to reduce experimental costs. Due to space constraints, the main results are summarized in the main text, with detailed dataset samples and additional results provided in Appendix~\ref{sec:0}.

\begin{figure*}[t]
    \centering
    \includegraphics[width=\textwidth]{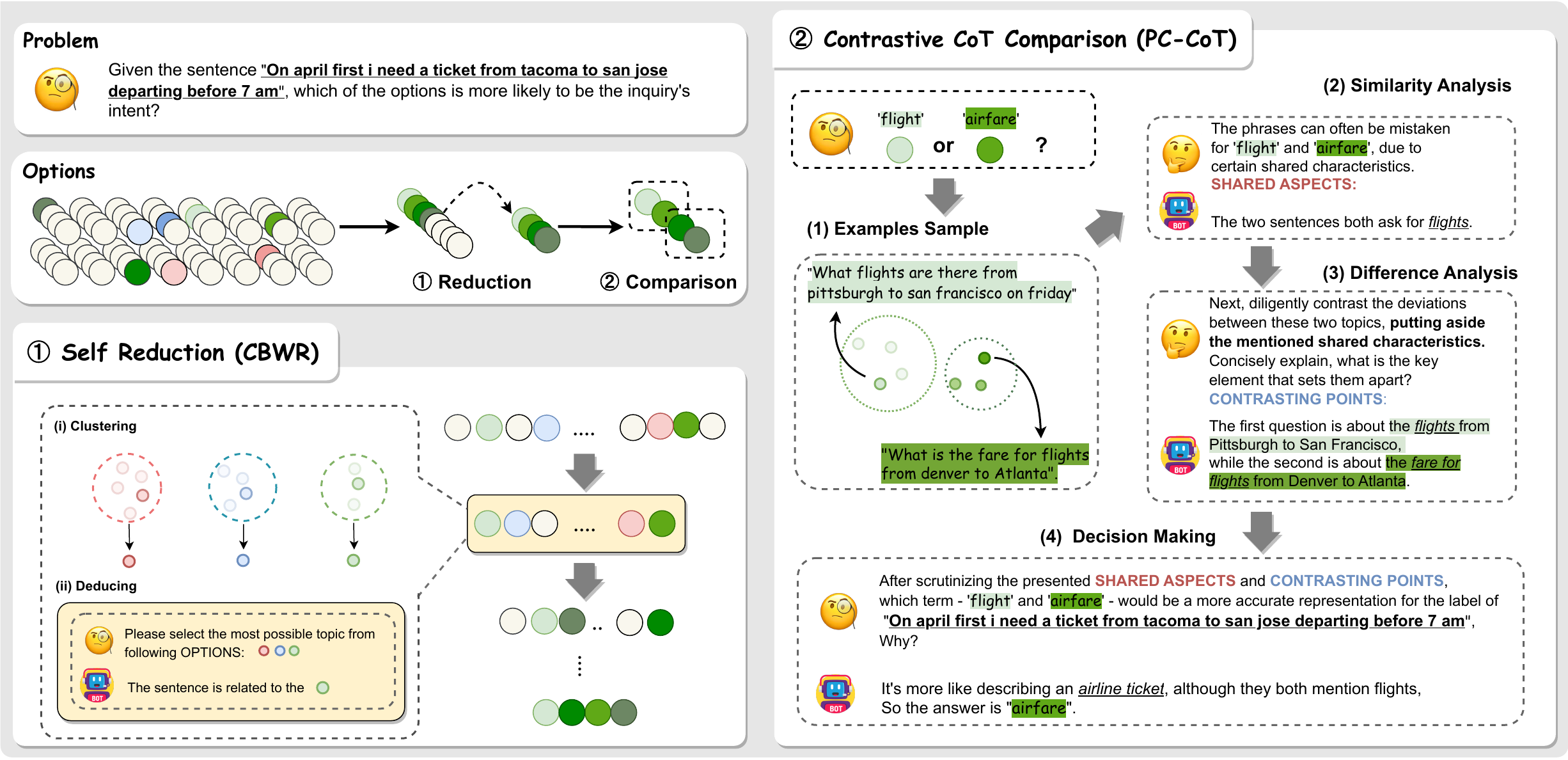}
    \caption{
        An overview of our framework for text classification with numerous options. It comprises two main stages: (1) Self Reduction (CBWR) which eliminates extraneous options, and (2) Contrastive CoT Comparison (PC-CoT) that conducts pairwise comparison.  
        % Our algorithm employs a multi-iteration self-reduction algorithm to isolate the most probable option. 
        In CBWR, distinct option ``windows'' are induced via clustering to avoid confusion due to similarity. 
        For pairwise comparison, our PC-CoT employs a contrastive chain-of-thought technique to highlight similarities and differences among the paired options for final decision making. }
    \label{fig:main}
\end{figure*}

\paragraph{Key Observations}

% \textcolor{red}{We conduct an extensive evaluation across various LLMs to establish a preliminary comprehension of factors influencing classification accuracy. }
Our findings, presented in Figure~\ref{fig:sim} and Figure~\ref{fig:3body}, reveal the following key observations and insights:

\textbf{(1) The option similarity consistently poses a challenge.}
% Similar options contribute to the creation of ambiguous decision boundaries within these models. 
% In a straightforward test involving 16 randomly sampled or similarity-retrieved options, 
We test randomly sampled or similarity-retrieved options in Figure~\ref{fig:sim} and observe a consistent performance drop across various LLM sizes and types. 
This finding indicates that LLMs encounter difficulties when faced with the inherent fuzziness of semantics. 
% This underscores the need for further investigation into the nuanced challenges associated with option ambiguity.

\textbf{(2) The increased number of options consistently presents a challenge for LLMs.} 
Across various LLMs, including LLaMA, Qwen, and gpt-3.5-turbo, we consistently observe a decline in performance as the number of options increases, as shown in Figure~\ref{fig:a}. 
% Furthermore, different LLMs exhibit varying degrees of performance downgradations in response to the increased number of options, with the gpt-3.5-turbo series demonstrating heightened stability. 
{In addition, this challenge persists when a longer context window is introduced, \textit{e.g.,} the gpt-3.5-turbo-1106 with 16k context length and LongAlpaca with 32k context length.} 
% though they may show a more gradual performance decline. 
This phenomenon suggests that the noise stemming from the proliferation of options remains a significant obstacle in the landscape of LLMs.

\textbf{(3) Bias towards option positions and tokens.} 
In Figure~\ref{fig:b}, we discern a significant bias caused by the position of the correct choice, especially in open-source models such as LLaMA and Qwen, resulting in performance fluctuations exceeding 50\%. 
In contrast, gpt-3.5-turbo demonstrates minimal effects, with fluctuations below 10\%. 
% highlighting its superiority over open-source counterparts.
This lack of robustness demonstrates inherent position bias in LLMs.
% as elucidated by ~\citet{pezeshkpour2023large}.
% The bias in option positions becomes particularly apparent when the span of option choices is extensive, as the prolonged context amplifies the imbalance in position, presenting challenges for maintaining stable performance in classification problems.
In Figure~\ref{fig:c}, we analyze the confusion matrix of LLaMA for the LIU54 benchmark. A preference for ``movies'' and ``convert'' tokens can be clearly observed regardless of the golden label, which indicates the existence of token bias.

\textbf{{(4) The above issues cannot be effectively mitigated by long-context capabilities, simple prompting strategies, or single increasing the model size.}}
Even with a longer context, issues persist regarding the number and placement of options. For example, the gpt-3.5-turbo-1106 (16k context length) exhibits similar declining trends as its 4k-context counterpart with increasing option numbers.
The LongAlpaca-32k has 50\% performance fluctuation when the golden option is fixed at the $50^{th}$ position. 
Furthermore, experiments with varying numbers of options show consistent trends in both few-shot and zero-shot scenarios. There are distinct bias distributions between 3-shot and zero-shot scenarios, indicating that in-context examples can alter the bias distributions of option positions but cannot effectively eliminate them.
Regarding the model size, we observe that LLMs of different sizes all suffer from option similarity, as shown in Figure~\ref{fig:sim}. Even 70B LLaMA and Qwen models suffer from an increase in option number and simple arrangement, as depicted in Figure~\ref{fig:3body}.

\paragraph{From Full Option to Pairwise Comparison}

Inspired by {Figure~\ref{fig:a}, selecting the correct label from two choices} is much easier for LLMs than directly selecting from all options. For instance, gpt-3.5-turbo achieves a 94.29\% zero-shot accuracy when choosing between two categories. To this end, we propose a pairwise comparison framework to maximize the capabilities of LLMs. Our approach focuses on comparing pairs of options, which reduces the range of positions involved in the classification process. This reduction has been shown to effectively mitigate position bias~\citep{pezeshkpour2023large}. Furthermore, using pairwise comparisons allows for more nuanced comparative strategies, avoiding token shortcuts and ensuring that the most distinguish content guides decision-making.

\section{Methodology}

\subsection{Formulation}

Formally, given a fixed LLM parameterized by ${\theta}$, we use $x$ to denote the text classification input, $\mathcal{Y}=\{y_1, y_2, \cdots, y_N\}$ as $N$ potential label options. 
Then, we utilize the prompt to instruct LLM to select the appropriate label $y \in \mathcal{Y}$ for input $x$: 
\begin{equation}
    \begin{aligned}
        y \sim p_{\theta}(y|\texttt{Prompt}(x, \mathcal{Y}))
    \end{aligned}
\end{equation}
For simplicity, we will subsequently refer to $p_{\theta}(y|\texttt{Prompt}(x, \mathcal{Y}))$ as $p_{\theta}^{\texttt{Prompt}}(y|x, \mathcal{Y})$

The expressions for in-context learning ($p_{\theta}^{\texttt{ICL}}$) and chain-of-thought methodology ($p_{\theta}^{\texttt{CoT}}$)  for tackling text classification problems are formulated as:
\begin{equation}
    \begin{aligned}
        &p_{\theta}^{\texttt{ICL}}(y \mid S_{icl}, x, \mathcal{Y}),\\
        &p_{\theta}^{\texttt{CoT}}(z,y \mid S_{cot}, x, \mathcal{Y})\\
    \end{aligned}
\end{equation}
where the set $S$ denotes the demonstration sets, with an empty size ($S = \varnothing$) in the zero-shot scenario and a non-empty size in the few-shot scenario. 
For the few-shot instances, $S_{icl}$ includes pairs from $\{x^{\prime},y^{\prime} \mid y^{\prime} \in \mathcal{Y}, x^{\prime} \neq x \}$. For CoT methods, $S_{cot}$ additionally include texts $z^{\prime}$, viewed as the step-by-step reasoning ``thoughts''.
These thoughts serve as instructional support for LLMs to induce the conditional distribution of $y^{\prime}$ based on the given $x^{\prime}$.

\subsection{Proposed Framework}

The fundamental concept of our approach is to compare options in a {pairwise} manner.
% This approach is engineered to address obstacles associated with ambiguity and bias.
In cases where the options set is substantial, directly contrasting original options in pairs becomes inefficient and costly. Therefore, we propose a two-stage framework:
(i) \textbf{Reduction}, strategically eliminating extraneous options $\mathcal{Y}$ to a reduced set $\mathcal{R} \subset \mathcal{Y}$.
(ii) \textbf{Comparison}, navigating through the interference options within $\mathcal{R}$ via pairwise comparison in a chain-of-thought fashion. 
% The core idea of our method is guiding LLM to evaluate options by comparing them in a \textbf{pairwise} manner,
% This approach is engineered to address obstacles associated with ambiguity and bias.
% In cases where the options set is substantial, directly contrasting original options in pairs becomes inefficient and costly. Therefore, we propose a two-stage framework:
% (i) \textbf{Reduction}, strategically eliminating extraneous options $\mathcal{Y}$ to a reduced set $\mathcal{R} \subset \mathcal{Y}$.
% (ii) \textbf{Comparison}, navigating through the interference options within $\mathcal{R}$ via pairwise comparison in a chain-of-thought fashion. 
% The overall framework is succinctly summarized in Equation~\ref{eq:overall}:
% \begin{equation}
%     \begin{aligned}
%         &p_{\theta}\left(z, y \mid S, x, \mathcal{Y} \right)\\
%         =& \underbrace{p_{\theta}\left(\mathcal{R} \mid x, \mathcal{Y} \right)}_{\text{(i) Reduction}} \cdot \underbrace{p_{\theta}\left(z, y \mid S, x, \mathcal{R} \right)}_{{\text{(ii) Comparison}}}
%     \end{aligned}
%     \label{eq:overall}
% \end{equation}
In our two-stage framework, various reduction and comparison algorithms can be seamlessly integrated. This paper further introduces two new algorithms designed for this framework: self-reduction (in Section \ref{sec:3.3}), which utilizes LLMs to filter out irrelevant options automatically, and the contrastive chain-of-thought technique (in Section \ref{sec:3.4}), which mimics how humans compare confusable objects. The overall architecture of our algorithm is illustrated in Figure~\ref{fig:main}.

\subsection{Self Reduction}
\label{sec:3.3}

Given a sentence $x$ and option sets $\mathcal{Y}$, LLMs iteratively refine the options to $\mathcal{R}$ through a multi-iteration strategy, \textit{i.e.,} $p_{\theta}\left(\mathcal{R} \mid x, \mathcal{Y} \right)$.
Specifically, we design two types of multi-iteration strategies to improve the elimination accuracy: 
(1) Iterative Top Reduction (ITR): 
where LLMs iteratively choose the most likely options from the given candidate set and subsequently isolate them in the following iteration. 
This cyclic process continues until a pre-set threshold is reached, facilitating the steady refinement of labels.
(2) Cluster-Based Window Reduction (CBWR): 
Inspired from Figure~\ref{fig:sim}, a high degree of option similarity may degrade selection precision.
To mitigate this, CBWR refines the full set of options for an input sentence through iterative clustering and selection within a window. The remaining options are clustered into $K$ clusters, with samples evenly sampled from each cluster and placed into a window. Top-N candidates are then selected from the window by LLM, progressively discarding less probable options.
The detailed procedure can be found in Algorithm~\ref{alg:cbwr} in Appendix~\ref{app:a}.

% These strategies align with previous recommendations~\citep{pezeshkpour2023large}, emphasizing that organizing top choices among the adjacent options can reduce position bias.
% Because after the self-reduction process, the top choices naturally cluster together. 

\subsection{Contrastive CoT Comparison}
\label{sec:3.4}

In this phase, the remaining candidates within the set $\mathcal{R}$ often exhibit significant ambiguity.
% Each option is likely to have a close relationship with input $x$ from the perspective of LLMs or might be confusable with other options, 
% posing challenges for accurate classification. 
Each option has a close relationship with input $x$ from the perspective of LLMs. 
To accurately identify the candidate, we propose a novel strategy based on a ``similar-difference-decide'' reasoning chain during the pairwise comparison process, which we termed PC-CoT.
This approach is inspired by the concept of contrastive learning. 
% which takes both shared and distinct perspectives of features as informative supervisory signals. 

As specified in Algorithm~\ref{alg:maccot}, our approach begins by initiating a candidate pool with the reduced label set, systematically selecting two labels in each iteration, denoted as $y_1, y_2$. 
Subsequently, the demonstration set expands to include the demonstrations of both labels, \textit{i.e.,} $S = S_{y_1} \cup S_{y_2}$. 
The LLMs are then prompted to analyze the similarity ($z_s$) and differences ($z_d$) of demonstration sentences associated with these labels.
This step aims for a better understanding of the relationship between the two candidates and the refinement of the final decision on which label to eliminate. 
This iterative removal of candidates continues until we arrive at a sole label, which is regarded as the final answer.
The algorithm exhibits a time complexity of $O(|\mathcal{R}|)$. 
Given that the size of $\mathcal{R}$ is typically small, our approach ensures a cost-efficient performance.

% \begin{equation} \small
% \begin{aligned}
%     & p_{\theta}\left(z, y \mid \mathcal{R} \right) 
% = \prod_{{\begin{subarray}{c}y_1, y_2 \in \mathcal{R},\\ y_1 \neq y_2\end{subarray}}} p_{\theta}\left(z, y \mid \{y_1, y_2\} \right) \\ 
% & = \prod_{{\begin{subarray}{c}y_1, y_2 \in \mathcal{R},\\ y_1 \neq y_2\end{subarray}}} 
% \underbrace{p_{\theta}^{\texttt{Sim}} \left(z_{s} \mid \{y_1, y_2\}  \right)}_{\text{similarity analysis}} \cdot \\
% & \underbrace{p_{\theta}^{\texttt{Diff}} \left(z_{d} \mid  z_{s},\{y_1, y_2\}  \right)}_{\text{difference analysis}} \cdot 
% \underbrace{p_{\theta}^{\texttt{Dec}} \left( y \mid z_{s},z_{d}, \{y_1, y_2\} \right)}_{\text{decision-making}}
% \end{aligned}
% \label{eq:n2}
% \end{equation}

\input{algos/compair.tex}

\input{tables/reducer.tex}

\input{tables/compair.tex}

\section{Experiments}
\label{sec.4}

To evaluate the performance of our framework, 
we choose four widely recognized classification datasets, including:
\textbf{Banking77} \citep{casanueva2020efficient} with 77 labels in the banking services domain;
\textbf{HWU64} \citep{casanueva2020efficient} of 64 intents from 21 different domains; 
\textbf{LIU54} \citep{liu2019benchmarking} of 54 classes gathered from Amazon Mechanical Turk;
\textbf{Clinc150} \citep{larson2019evaluation} covering 150 classes and 10 domains.
Notice that we sample challenging segments from these datasets to reduce the experiment cost. The comprehensive sampling process and supervised performance are detailed in Appendix~\ref{app:samp}.
Our experiments primarily utilized gpt-3.5-turbo alongside open-source models like LLaMA2-Chat and Qwen-Chat series. Here we use the instruction versions of models for better instruction following ability.  
% over pre-trained ones as pre-trained models are used for completion and introduce inaccuracies in extracting answers.
% \citep{touvron2023llama2} and Qwen-Chat series \citep{bai2023qwen}. 
We set the temperature to 0 for all models across different experiments.
% For open source models, We mainly report the results from their 70B versions, with additional size-related outcomes in Appendix~\ref{app:os}.

\subsection{Evaluation For the Reduction Stage}
\paragraph{Setups.}
For evaluating the performance of the reduction stage and our self-reduction algorithm, 
we configure the size of the reduced set as $|\mathcal{R}|=5$ and report the HIT@5 metric on the benchmark.
This metric measures the proportion of instances where the correct label is among the top 5 labels chosen by the label reducer.
We compare the effectiveness of our two techniques, ITR and CBWR, against the following baselines:
(1) standard prompt, manually prompts the LLM to select the top 5 labels in a single call. 
(2) self-consistency~\citep{wang2023selfconsistency}, which invokes the aforementioned prompt multiple times, followed by a voting process to determine the final results. 
Appendix~\ref{app:a} provides additional details about their prompts.

\paragraph{Results.}
Table~\ref{tab:red} provides a detailed comparison of HIT@5 performance. 
It is obvious that both self-reduction strategies ITR and CBWR consistently outperform the standard prompt and the self-consistency method by a relatively large margin. 
The self-consistency approach also exhibits an advantage over the standard prompt for gpt-3.5-turbo and Qwen-72B.
However, it may degrade the performance of the LLaMA2-70B model, which could result from the differences in training data.
Another observation is significantly low reduction performance occurs on the LIU54 dataset. 
This could be attributed to a large proportion of short sentences in LIU54, making them challenging to interpret and reason effectively, more details are listed in Appendix~\ref{app:liu}.
Despite the above challenges, our strategies can consistently improve the reduction performance, minimizing error propagation to the next stage.      

\subsection{Evaluation For Overall Framework}
\paragraph{Setups}
We primarily use accuracy as the performance metric for our framework, considering both the reduction and pairwise comparison stages, 
The selected baselines include common prompting techniques: zero-shot (ZS), zero-shot-CoT (ZS-CoT), few-shot (FS), and few-shot-CoT (FS-CoT). 
We evaluate each method in two scenarios: with and without the reduction stage, to study the impact of the overall framework and compare the pairwise comparison algorithm. 
In the former scenario, we prompt the LLM directly with the full set of options, while in the latter, we conduct comparison experiments based on the CBWR result for gpt-3.5-turbo and Qwen, and the ITR result for LLaMA2.
It is important to note that for the latter scenario, we compare four prompting methods and our proposed PC-CoT using the same reduction technique, to ensure a fair evaluation for the comparison stage. 
For the few-shot method, if the text exceeds the context length of 4096, we randomly omit demonstrations to reduce the context length. 
To avoid random errors, we conduct the experiments on each dataset five times and present the results in terms of mean and standard variance. 
Appendix~\ref{app:b} provides further details regarding the prompts.

\paragraph{Results}
As shown in Table~\ref{tab:com}, we have the following observations: 

(1) Our framework significantly improves the stability and predictability of option selection in ZS, ZS-CoT, FS, and FS-CoT settings. 
Specifically, it boosts the zero-shot performance of gpt-3.5-turbo relatively by 11.88\%, the few-shot performance of LLaMA2 from 26.45\% to 54.01\% (see Section~\ref{sec:debias} for detail analysis), and the FS-CoT performance of Qwen is relatively improved by 10.95\%.
These results demonstrate the effectiveness of our framework in mitigating ambiguity and bias.

(2) Our PC-CoT paired with our framework stands out with the best results for gpt-3.5-turbo and Qwen. Notably, it achieves a relative increase of 54.1\% in average accuracy over a directly selected answer for gpt-3.5-turbo, from 40.24\% to 62.01\%.  
However, for LLaMA2-70B, our performance is slightly lower than the FS method on the Banking77 and Clinc150 benchmarks. We hypothesize this may have some relationship with LLaMA2 model training data, which makes it insensitive to the chain-of-thought guidance related to classification problems because we can observe that FS-CoT demonstrates a degradation in performance, dropping from 57.55\% under FS to 39.18\%.

(3) Compared to original CoT techniques, our PC-CoT demonstrates better generality across different types and sizes of LLMs. 
While the FS-CoT method exhibits satisfactory performance with the Qwen model, it deteriorates the performance of LLaMA and even hinders the zero-shot performance of gpt-3.5-turbo on Clinc150 and LIU54.
This result suggests that the prompting method in classification scenarios requires tailored adjustments to effectively prioritize comparison for ambiguous boundaries.

The detailed cases of all nine prompting methods in Table 2 can be found in Appendix~\ref{app:case}.

\subsection{Ablation Study of PC-CoT algorithm}

% \paragraph{The Effectiveness of Our Overall Framework}
% \paragraph{The Effectiveness of Our Propose PC-CoT algorithm}
\input{tables/ablation.tex}
As outlined in Table~\ref{tab:ab}, we demonstrate the impact of removing key components from our PC-CoT. 
Without the similarity or difference analysis prompt, we observe an evident performance decline (approximately 15\%). Furthermore, using only the difference analysis prompt leads to inferior results, 
supporting our belief that we need to discard the confusable points to better extract the distinctive features effectively.
When both similarity and difference analyses are removed, our method reverts to plain FS-CoT, thus exhibiting a performance level similar to FS-CoT.

\begin{figure}[t]
    \centering
    \begin{subfigure}[h]{0.23\textwidth}
        \centering
        \includegraphics[width=\textwidth]{figures/llama_fewshot_liu54.json.pdf}
        \captionsetup{justification=centering}
        \caption{FS-LIU54 \\(Full Options)}
    \end{subfigure}
    \begin{subfigure}[h]{0.23\textwidth}
        \centering
        \includegraphics[width=\textwidth]{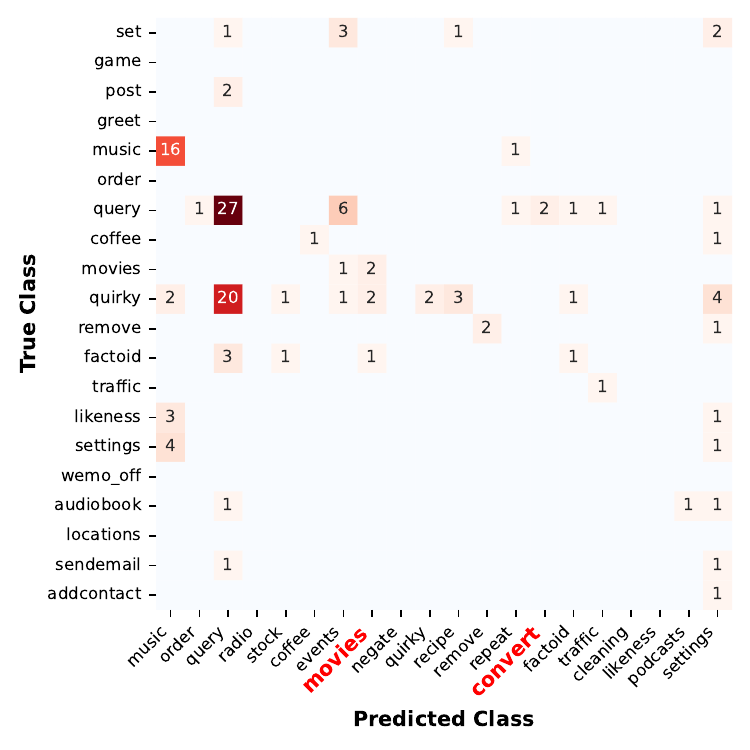}
        \captionsetup{justification=centering}
        \caption{FS-LIU54 \\(Our framework)}
    \end{subfigure}
    \begin{subfigure}[h]{0.23\textwidth}
        \centering
        \includegraphics[width=\textwidth]{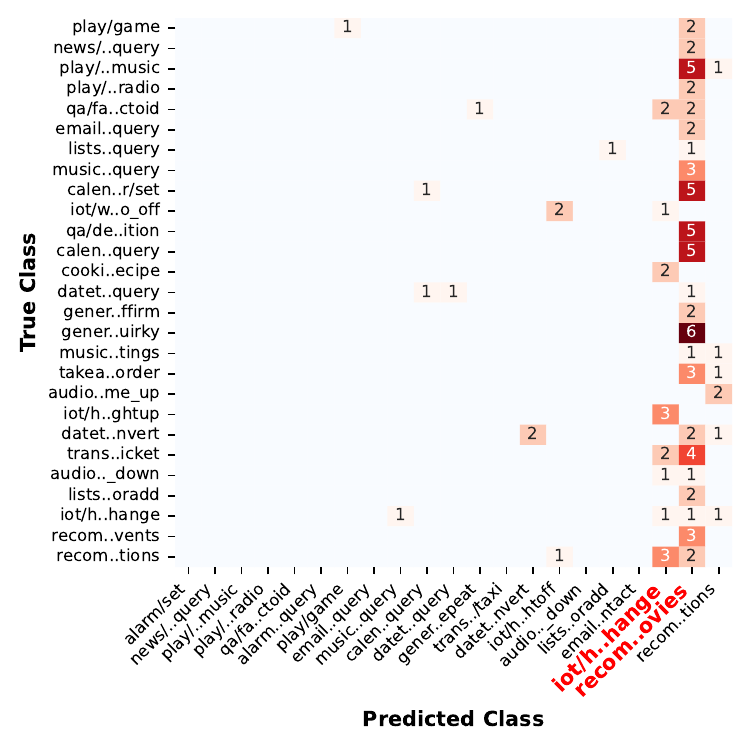}
        \captionsetup{justification=centering}
        \caption{ZS-CoT-HWU64 \\(Full Options)}
    \end{subfigure}
    \begin{subfigure}[h]{0.23\textwidth}
        \centering
        \includegraphics[width=\textwidth]{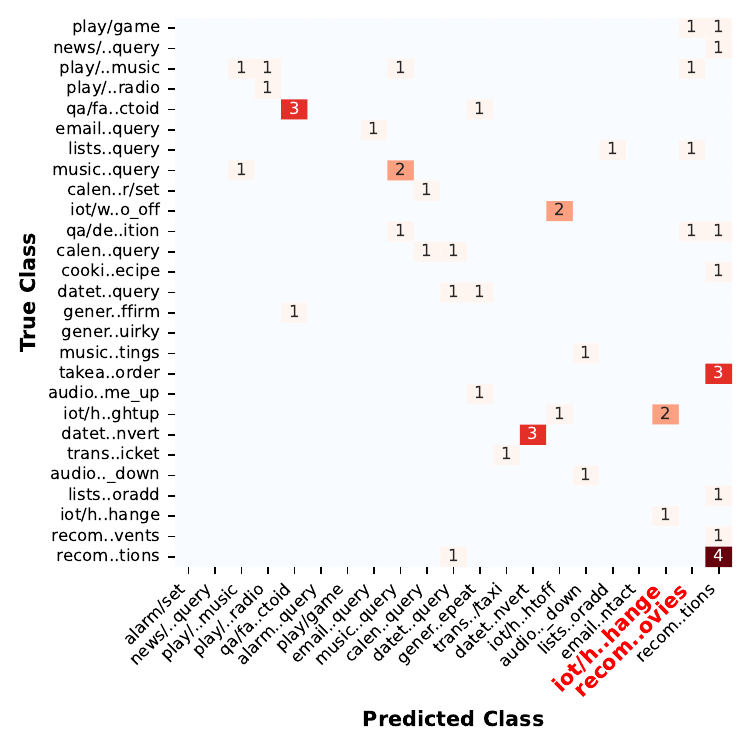}
        \captionsetup{justification=centering}
        \caption{ZS-CoT-HWU64 \\(Our framework)}
    \end{subfigure}
    \caption{The confusion matrix visualization illustrates the debiasing effect of our framework on LLaMA70B. In the original few-shot setting, the model showed a preference for ``movies'' and ``convert'', while in the ZS-CoT setting, it preferred ``iot/hue\_lightchange'' and ``recommendation/movies''. After applying our framework, the preference bias clearly shifted. }
    \label{fig:debias}
\end{figure}

\begin{table*}[t]
\centering
\resizebox{0.9\linewidth}{!}{
\begin{tabular}{lccccccc}
    \toprule
    \textbf{Method} & \textbf{Full Options FS} & \textbf{P + FS} & \textbf{SC + FS} & \textbf{SC + FS-CoT} & \textbf{ITR + FS} & \textbf{CBWR + FS} & \textbf{CBWR+PC-CoT} \\
    \midrule
   \textbf{ \#Call Avg.}           & 1              & 5      & 12      & 24          & 9        & 8         & 12          \\
    \textbf{Time / 1000 items} & 2min2s           & 11min32s & 28min1s & 53min1s     & 18min55s & 14min21s  & 25min56s    \\
    \textbf{Performance Gain }     & 0              & -0.39  & +0.21   & +0.19       & +3.89    & +5.01     & +8.74       \\
    \bottomrule
\end{tabular}
}
\caption{ LLM Calls Comparison}
\label{tab:reduction_comparison}
\vspace{-5mm}
\end{table*}

\begin{table}[t]
    \centering
        \resizebox{\columnwidth}{!}{
    \begin{tabular}{lcccc}
        \toprule
        \textbf{Method} & \textbf{Banking77} & \textbf{HWU64} & \textbf{LIU54} & \textbf{Clinc150} \\
        \midrule
        \textbf{Retrieval-Augmented ICL} & 56.25 & 52.66 & 35.2 & 78.73 \\
        \textbf{Ours} & 61.83 & 65.09 & 37.35 & 83.78 \\
        \bottomrule
    \end{tabular}
    }
    \caption{Comparison with in-context learning baselines}
    \label{tab:a}
    \vspace{-5mm}
\end{table}

\begin{table}[t]
    \centering
        \resizebox{0.9\columnwidth}{!}{
    \begin{tabular}{lcccc}
    \toprule
\textbf{Method} & \textbf{reminder} & \textbf{agnews} & \textbf{bbcnews} & \textbf{restaurant} \\ \midrule
\textbf{FS-CoT} & 78.79    & 81.31  & 92.19   & 86.98    \\
\textbf{Ours}   & 79.69    & 82.83  & 92.25   & 87.37    \\
        \bottomrule
    \end{tabular}
    }
    \caption{Performance on simple text calssification tasks}
    \label{tab:b}
\vspace{-5mm}
\end{table}

\begin{table*}[t]
\centering
\resizebox{0.9\linewidth}{!}{
\begin{tabular}{llcccccccccc}
    \toprule
    \multicolumn{1}{c}{\multirow{2}{*}{\textbf{Model}}} & \multicolumn{1}{c}{\multirow{2}{*}{\textbf{Dataset}}} & \multicolumn{4}{c}{{\textbf{Full Options}}}  & \multicolumn{5}{c}{{\textbf{Our Framework}}} \\
    \cmidrule(r){3-6} \cmidrule(l){7-11}
    & & \textbf{ZS} & \textbf{ZS-CoT} & \textbf{FS} & \textbf{FS-CoT} & \textbf{ZS} & \textbf{ZS-CoT} & \textbf{FS} & \textbf{FS-CoT} & \textbf{PC-CoT} \\
    \midrule
    \multicolumn{1}{c}{\multirow{4}{*}{\textbf{Qwen-1.8B-Chat}}} & \textbf{Banking77} & 18.18 & 17.53 & 4.87 & 24.68 & 26.62 & 23.7 & 19.16 & \textbf{31.82} & 19.16 \\
    & \textbf{HWU64} & 21.82 & 22.73 & 4.55 & 15.45 & 24.55 & 26.36 & 20.04 & \textbf{29.09} & 23.64 \\
    & \textbf{LIU54} & 17.32 & 17.32 & 0.79 & 18.11 & 23.62 & \textbf{20.47} & 17.72 & 20.08 & 17.32 \\
    & \textbf{Clinc150} & 30.22 & 33.78 & 5.78 & 20.89 & 39.11 & 33.78 & 35.56 & \textbf{54.22} & 25.78 \\ \midrule
    \multicolumn{1}{c}{\multirow{4}{*}{\textbf{Qwen-14B-Chat}}} & \textbf{Banking77} & 33.44 & 33.12 & 38.64 & 46.43 & 35.06 & 26.62 & 51.95 & 46.10 & \textbf{56.82} \\
    & \textbf{HWU64} & 40.01 & 41.82 & 47.27 & 50.02 & 37.27 & 38.18 & 55.45 & 53.64 & \textbf{62.73} \\
    & \textbf{LIU54} & 29.53 & 27.95 & 31.50 & 27.95 & 31.10 & 25.98 & 37.40 & 32.28 & \textbf{38.19} \\
    & \textbf{Clinc150} & 53.78 & 52.44 & 58.22 & 67.11 & 55.56 & 42.22 & 77.78 & 73.78 & \textbf{82.67} \\
    \bottomrule
\end{tabular}
}
\caption{Performance Comparison of LightWeight LLMs}
\label{tab:c}
\end{table*}

\subsection{Debias Study}
\label{sec:debias}

We find that our framework significantly improves the ZS-CoT and FS performance of LLaMA2 in Table~\ref{tab:com}. 
This enhancement is particularly evident in the reduced bias effect of certain tokens, such as ``movies'' and ``convert'' in the LIU54 dataset, and ``iot/hue\_lightchange'' and ``recommendation/movies'' in the HWU64 dataset. 
By visualizing the confusion matrices for both datasets (Figure~\ref{fig:debias}), we can observe a noticeable shift in preferences after applying our framework, demonstrating its effectiveness. 
This improvement can be attributed to our reduction technique and pairwise comparison approach, which reduce the number of tokens and positions considered, thereby mitigating noise interference and enhancing the capabilities of few-shot demonstration and CoT reasoning.

\subsection{Efficiency Analysis}

Our method shows promising results but may require more LLM calls due to multiple iterations for reduction and pairwise comparisons to enhance accuracy and reduce bias. To assess the applicability of our approach, we compare our inference speed and call frequency with other methods in Table \ref{tab:reduction_comparison}. Note that all baseline methods require multiple calls to LLMs, except for partial methods using the full option. However, the full option struggles with boundary ambiguity and inherent biases, leading to low accuracy and practical deployment difficulties. Our approach significantly outperforms other baselines by adding only a limited number of calls and achieves the highest performance gain (as shown in the last line of Table \ref{tab:reduction_comparison}). In contrast, self-consistency-based methods require more calls than our approach but offer limited performance improvements. With the assistance of technologies such as flash-attn and vLLM, we have successfully reduced the computational overhead of our method when calling the 70B model. As a result, we can accurately predict 1,000 pieces of data in just half an hour, which is significantly faster than the multiple-call methods used in Self-Consistency. Moreover, the future of technology is headed towards large-scale models. It is foreseeable that with technological advancements, the costs associated with API calls and model computations will decrease further. Our method, with its enhancements addressing bias and boundary issues, will become increasingly relevant and significant in solving real-world classification problems. Therefore, we believe that the scaling trend of language models will elevate the importance of our method in the future.

\subsection{In Context Learning}

We experiment with a similar in-context learning baseline \cite{milios-etal-2023-context} that also uses clustering algorithms in a few-shot setting. As shown in Table \ref{tab:a}, our approach consistently outperforms the baseline. The main reason is that clustering similar demonstrations can blur the model's decision boundary when options are highly confusable, reducing performance. In contrast, our method introduces diversity through ITR/CBWR and provides a more detailed comparison via PC-CoT.

\subsection{Simple Text Classification Tasks}

Our framework is appliable to simple text classification tasks and has comparable performance to the few shot chain-of-thought method, as shown in Table \ref{tab:b}. In our primary experiments, we focused on classification tasks with over 50 categories to demonstrate our framework's ability to handle unclear decision boundaries arising from numerous options and similar, easily confused elements.

\subsection{Lightweight Models}

As shown in Table \ref{tab:c}, our method still applies to smaller models. While the 14B model outperforms other baselines, the Qwen 1.8B model slightly underperforms compared to the Few-Shot CoT methods. Because smaller models (<10B) may struggle with tasks such as instruction following and generating high-quality few-shot examples, as highlighted in prior research \cite{pmlr-v202-fu23d}.

\section{Related Work}

\paragraph{Text Classification}
Traditional text classification approaches typically involve feature extraction by RNN~\citep{xie2020unsupervised}, GCN~\citep{lin2022bertgcn}, or Transformers~\citep{chai2020description}.
These features are subsequently input into a classifier head and assigned the corresponding label. 
To deal with the common data scarcity issue, the meta-learning paradigm emerges to identify novel categories with minimal labeled examples by leveraging prior knowledge acquired from known categories\citep{2023ContrastNetContrastiveLearning,gao2021making,2023EffectiveStructuredPrompting,2023BoostingFewShotText,2023PretrainingIntentAwareEncoders}.
Nonetheless, these approaches require supervised fine-tuning and exhibit limited applicability.
The powerful capabilities of modern LLMs enable zero-shot or few-shot classification without necessitating model update ~\citep{2023ReasoningImplicitSentiment,2023BreakingBankChatGPT}.
% This approach, particularly when coupled with prompting techniques like in-context learning~\citep{brown2020language} and chain-of-thought~\citep{wei2023chainofthought}, has gained prominence. 
Despite these advancements, the exploration of LLM-based classification methodologies in addressing practical challenges, such as coping with ambiguous boundaries and inherent bias, remains underexplored.

\paragraph{Chain of Thought}
\label{sec2-2}
\citet{wei2023chainofthought} introduces few-shot chain-of-thought, 
enriching LLM reasoning capabilities by incorporating intermediate reasoning steps in each demonstration.
\citet{kojima2023large} highlights the efficacy of a straightforward prompt, ``Let's think step by step'', in enhancing zero-shot reasoning.
While recent investigations have explored diversifying demonstrations~\citep{2022AutomaticChainThought} or modeling the reasoning process using tree and graph structures~\citep{yao2023tree,lei2023boosting},
the realm of classification problems continues to lack specialized CoT methods to combat ambiguity and bias. 
THOR \citep{2023ReasoningImplicitSentiment} exclusively addresses sentiment analysis, 
sequentially inducing aspects and opinions for sentiment identification. 
CARP~\citep{sun-etal-2023-text} focuses on simple classification problems with only 2-4 options and utilizes supervised fine-tuning of the model, which is impractical in more complex classification scenarios.
In contrast, we focus on handling a large number of label options and subtle semantic differences. We also address the migration of preferences for tokens and positions, enabling seamless integration into a wide range of text classification tasks.

% Text classification, a fundamental task that assigns predefined labels to a given text, has a wide spetrum of applications.  
%or the inherent option position bias in LLMs.
% To deal with the common data scarcity issue in real-world classification, the meta-learning paradigm emerges as a solution. 
% This paradigm seeks to identify novel categories with minimal labeled examples by leveraging prior knowledge acquired from known categories. 
% Diverse methodologies, including multi-level contrastive losses \citep{2023ContrastNetContrastiveLearning}, prompt learning \citep{schick2021exploiting,gao2021making,2023EffectiveStructuredPrompting}, data augmentation \citep{2023BoostingFewShotText} and encoder pretraining \citep{2023PretrainingIntentAwareEncoders}, have been explored within this context. 
% \textit{e.g.}, binary sentiment analysis~\citep{pontiki-etal-2014-semeval} or four-label topic classification~\citep{zhang2016characterlevel}.

% Auto-CoT~\citep{2022AutomaticChainThought} finds the diversity of few-shot demonstrations can enhance model performance. 
% The significant advancements of large-scale language models, such as ChatGPT\footnote{\url{http://chat.openai.com/}} and LLAMA~\citep{touvron2023llama,touvron2023llama2}, in showcasing human-level intelligence, 

\section{Conclusions}

In this paper, we investigate the vulnerability of LLMs in terms of option quantity and arrangement for text classification tasks. 
Our empirical analysis reveals that the main challenges lie in their ambiguous decision boundaries and inherent biases towards tokens and positions. 
To mitigate these challenges, we propose a novel two-stage classification framework for LLMs, involving reduction and comparison stages.
This framework strategically employs pairwise comparisons to address ambiguity and bias. Additionally, we introduce innovative self-reduction algorithms, including ITR and CBWR strategies, and our PC-CoT, a contrastive chain-of-thought technique, to enhance the framework's utility. 
Extensive experiments demonstrate our framework can bring improvement to the stability and debiasing for various prompting techniques. Our ITR and CBWR methods prove highly effective for reduction, while our PC-CoT exhibits superior performance in classification accuracy.

\section*{Limitations}

Though our method significantly outperforms other baselines by adding only a limited number of calls and achieves the highest performance gain, it does increase the number of LLM calls.
For some smaller-scale LLMs, their limited reasoning ability can impede precise reduction and comparisons, potentially diminishing the effectiveness of our method.

\section*{Ethics Statement}
In this study, the classification datasets and LLM responses used in our experiments only serve the specific purposes of evaluating our proposed approach.
To address ethical considerations, we have incorporated a rule-based filter during our challenge set sampling process to exclude offensive and sensitive information.
Our study investigates position bias and token bias during the classification process. 
Our algorithms help LLMs focus more on the actual content of options rather than making shortcut predictions based on preferences for specific tokens and positions, which is crucial for debiasing LLMs.

\section*{Acknowledgements}

This work was supported in part by the National Natural Science Foundation of China under Grant No. 62276110, No. 62172039 and in part by the fund of Joint Laboratory of HUST and Pingan Property \& Casualty Research (HPL). There are computational tasks are conducted using the HPC platform of Huazhong University of Science and Technology. The authors would also like to thank the anonymous reviewers for their comments on improving the quality of this paper.

\bibliography{tmp}
\bibliographystyle{acl_natbib}
\clearpage
\appendix
% \onecolumn

\begin{figure*}[htbp]
    \centering
    \begin{subfigure}[h]{\linewidth}
        \centering
        \includegraphics[width=\linewidth]{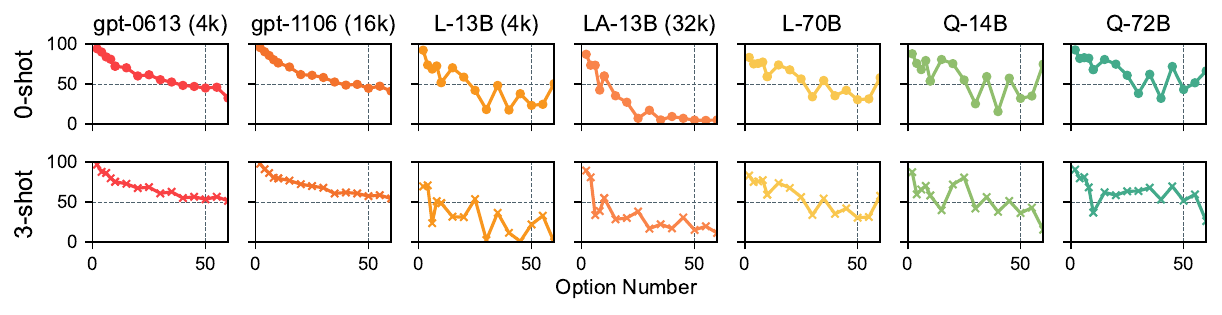}
        \caption{ LLMs classification accuracy concerning the number of options consistently demonstrates a decline. \label{fig:option_full}}
    \end{subfigure}
    \begin{subfigure}[h]{\linewidth}
        \centering
        \includegraphics[width=\linewidth]{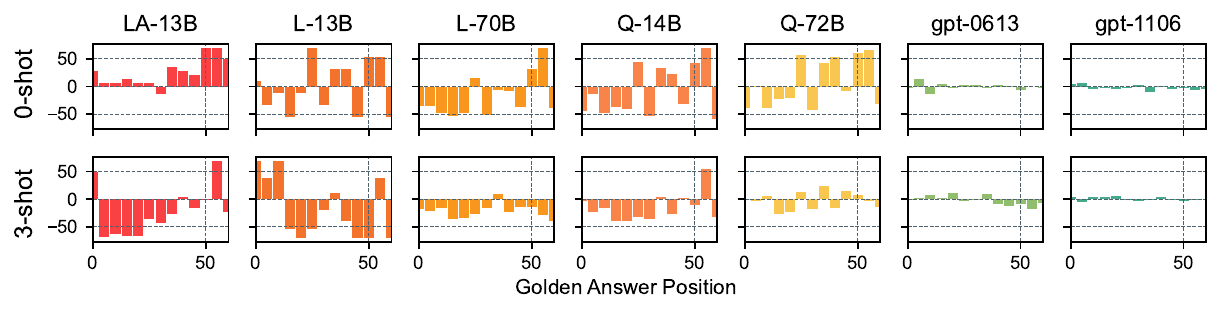}
        \caption{The change rate of classification accuracy after moving the correct option to a fixed position compared to the original accuracy. This simple option arrangement can lead to dramatic performance fluctuations in both zero-shot and few-shot scenarios. \label{fig:order_full}}
    \end{subfigure}
    \caption{The Full result for Figure~\ref{fig:a} and Figure~\ref{fig:b}
    }
    \label{fig:opt_full}
\end{figure*}

\section{More Detail of Preliminary Study And Dataset}
\label{sec:0}
\begin{figure*}[htbp]
    \centering
    % \begin{subfigure}[h]{0.49\textwidth}
    %     \centering
    %     \includegraphics[width=\textwidth]{figures/qwen18_fewshot_liu54.pdf}
    % \end{subfigure}
    % \begin{subfigure}[h]{0.49\textwidth}
    %     \centering
    %     \includegraphics[width=\textwidth]{figures/qwen7_zero_shot_cot_banking.json.pdf}
    % \end{subfigure}
    % \begin{subfigure}[h]{0.49\textwidth}
    %     \centering
    %     \includegraphics[width=\textwidth]{figures/llama70_zeroshotcot_hwu64.pdf}
    % \end{subfigure}
    % \begin{subfigure}[h]{0.6\textwidth}
        % \centering
        \includegraphics[width=\textwidth]{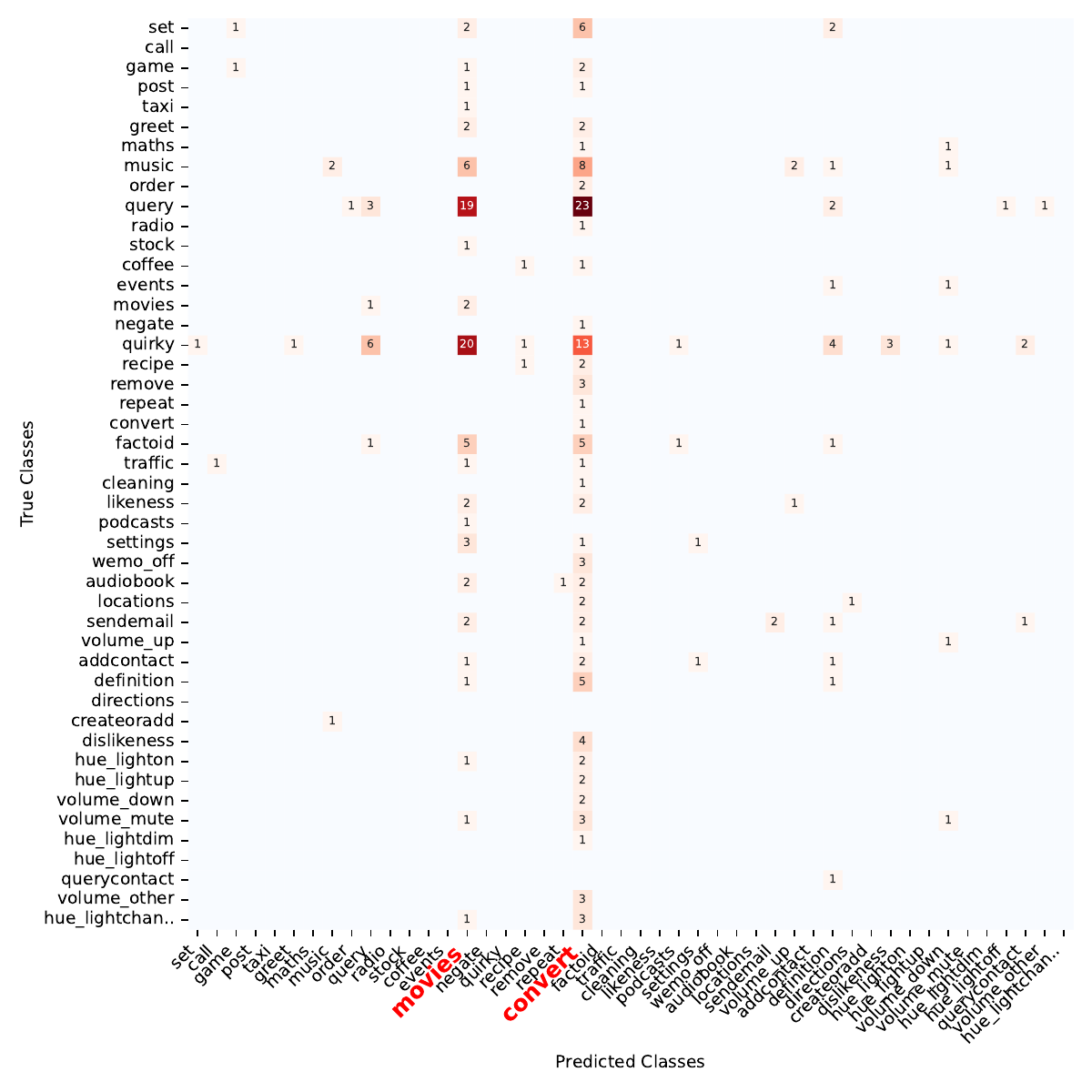}
    % \end{subfigure}
    
    \caption{Full confusion matrix from Figure~\ref{fig:c}.}
    \label{fig:bias_full}
\end{figure*}

\subsection{Full result for Figure~\ref{fig:3body}}

In Figure~\ref{fig:option_full}, we observe that across various LLM sizes and types, including LLaMA, Qwen, and gpt-3.5-turbo, we consistently observe a decline in performance as the number of options increases. 
Different LLMs may exhibit varying degrees of performance downgradations in response to the increased number of options, with the gpt-3.5-turbo series demonstrating heightened stability. 
The challenge persists even with longer contextual LLMs, though they may show a more gradual performance decline. 
This observation suggests that the noise stemming from the proliferation of options remains a significant obstacle in the landscape of LLMs.

In Figure~\ref{fig:order_full}, we discern a significant bias caused by the position of the correct choice, especially evident in open-source models such as LLaMA and Qwen, resulting in performance fluctuations exceeding 50\%. 
In contrast, gpt-3.5-turbo demonstrates minimal effects, with fluctuations below 10\%, highlighting its superiority over open-source counterparts.
This lack of robustness can be attributed to the inherent position bias inherent in LLMs, as elucidated by ~\citet{pezeshkpour2023large}.
The bias in option positions becomes particularly apparent when the span of option choices is extensive, as the prolonged context amplifies the imbalance in position, presenting challenges for maintaining stable performance in classification problems.

In Figure~\ref{fig:bias_full}, we show the full confusion matrix from Figure~\ref{fig:c}. A preference for ``movies'' and ``convert'' tokens can be clearly observed regardless of the golden label. Similar token biases are also observed for ``iot/hue\_lightchange'' and ``recommendation/movies'' for LLaMA on the HWU64 benchmark, or neglect the ``music'', ``query'', and ``quirky`` for the Qwen model.   

The long-context ability of LLM is insufficient in managing an increasing number of options or their respective positions.
Despite the common intuition that extended context should aid generative models in handling larger label sets, 
we find that performance diminishes with a growing number of labels, coupled with a persistent bias towards specific option positions. 
This is exemplified by comparisons such as gpt-3.5-turbo-1106 (16k context length) versus gpt-3.5-turbo-0613 (4k context length) and LongAlpaca (32k context length) versus LLaMA-13B (4k context length). 
Although a longer context does impart stability and smoothness in the face of increased options, 
challenges persist concerning both the number and positioning of options. the LongAlpaca-32k also suffers from 50\% performance fluctuation when changing the golden option to $50^{th}$ position.

The In-context examples may reduce option bias for large models while altering the distribution.
Examining in-context examples, we observe a reduction in bias for larger models like LLaMA-2-70B and Qwen-72B when moving from zero-shot to 3-shot demonstrations, while this effect is less pronounced for smaller models. 
Additionally, we note distinct bias distributions between 3-shot and zero-shot scenarios, suggesting that in-context examples may alter options position bias distributions.
In contrast, for experiments involving varying numbers of options, the trends observed in few-shot and zero-shot scenarios remain essentially consistent.

\subsection{Challenge Set Sampling}
\label{app:samp}

To align with real-world application scenarios, we employed a sampling strategy for a more challenging subset from these datasets (only their original test splits) by the precision margin~\citep{li2023instant}, 
$M(f(x)):= max(0, max_y f_y(x) - max_{k \neq y} f_k(x) $, which is an indicator of the confidence of the base classifier $f$ (the supervised fine-tuned BERT model).
\textit{i.e.,} the difference between the largest and second-largest label probability for every data item in the test dataset.
A smaller prediction margin correlates with decreased confidence in the prediction, representing a more challenging setting. 
As for the training dataset, we randomly sample five items from their training splits as our few-shot demonstrations. 

\subsection{Supervised Finetuning Performance on the challenge set}
\label{app:sft}

\input{tables/sft}
To better illustrate how challenging our curated dataset is, we select a series of encoder-only models and decoder-only LLMs for supervised fine-tuning (SFT) in a few-shot setting. 
For the encoder-only series, we fine-tune them for 30 epochs.
For the decoder-only series, due to the huge SFT cost, we only select TinyLLaMA-1.1B and Qwen-1.8B. We employ an alpaca format for fine-tuning (with the following prompt). The details of the results are listed in Table~\ref{tab:sft}. 

\begin{mybox}[FineTuning Prompt]
    \lstinline|Below is an instruction that describes a task. Write a response that appropriately completes the request.| \\

    \lstinline|### Instruction:|\\
    \lstinline|Given the sentence: "[{text}]", please select the most possible topic from the following OPTIONS: [{options}].| \\

    \lstinline|### Response: |\\
\end{mybox}

Table~\ref{tab:sft} indicates that the challenge datasets chosen pose significant difficulties for both BERT and RoBERTa, even when utilizing their larger versions and fine-tuning them for 30 epochs. With only five demonstrations, BERT achieves a mere 12.02\% accuracy, while RoBERTa fares slightly better at 23.70\%. In contrast, decoder-only language models, particularly Qwen-1.8B, demonstrate better performance than BERT and RoBERTa. However, this superior performance is contingent upon resource-intensive fine-tuning processes and may suffer from limited generalizability, thereby constraining their practical applicability.

\section{Detail of the Reduction Stage}
\label{app:a}

\begin{mybox}[Standard Reduction Prompt]
    \lstinline|Consider the sentence: "[{text}]"|\\
    \lstinline|Please select [{top_k}] most possible topic from following OPTIONS: [{options}] .| \\
    \lstinline|CHOICE:|
\end{mybox}

The self-consistency, our ITR, and CBWR techniques employ the same reduction prompt as exemplified above.
The specific procedural steps for CBWR are delineated in Algorithm~\ref{alg:cbwr}.  

\input{algos/reducer2.tex}

\section{Detail of the Comparison Stage}
\label{app:b}

\subsection{Baseline Using Full Options}

In this section, we list the prompt used in baselines without label reduction, \textit{i.e.,} the LLM is prompted to directly choose the correct label from the complete set of options.

\begin{mybox}[Zero-Shot Prompt ]
    \lstinline|Given the sentence: "[{text}]"|\\
    \lstinline|Please select the most possible topic from the following OPTIONS: [{options}]|\\
    \lstinline|CHOICE: |
\end{mybox}

\begin{mybox}[Zero-Shot-CoT Prompt ]
    \lstinline|Given the sentence: "[{text}]"|\\
    \lstinline|Please select the most possible topic from the following OPTIONS: [{options}]|\\
    \lstinline|Let's think step by step and give your explanation to verify your answer: |
\end{mybox}

\begin{mybox}[Few-Shot Prompt ]
    \lstinline|Below is a text classification problem, Note that you can only select the label in [{options}]|\\
    \lstinline|SENTENCE: [{text1}]|\\
    \lstinline|LABEL: [{label1}]|\\
    \\
    \lstinline|... |\\
    \\
    \lstinline|SENTENCE: [{text}]|\\
    \lstinline|LABEL:|
\end{mybox}

\begin{mybox}[Few-Shot-CoT Prompt  ]
    \lstinline|Below is a text classification problem, Note that you can only select the label in [{options}]. Let's think step by step and give your explanation to verify the answer.|\\
    \lstinline|SENTENCE: [{text1}]|\\
    \lstinline|EXPLANATION: [{explain1}]|\\
    \lstinline|LABEL: [{label1}]|\\
    \\
    \lstinline|... |\\
    \\
    \lstinline|SENTENCE: [{text}]|\\
    \lstinline|LABEL: [{label}]|\\
    \lstinline|EXPLANATION:|
\end{mybox}

Notice that we obtain the explanation demonstrations for Few-Shot-CoT by prompting LLM to automatically generate the explanation according to the golden label without human annotation. 
The prompt is as follows:

\begin{mybox}[Few-Shot-CoT Explanation Generation  ]
    \lstinline|Below is a text classification problem. Let's think step by step and give your explanation to verify the SENTENCE label:|\\
    \lstinline|SENTENCE: Fears for T N pension after talks Unions representing workers at Turner Newall say they are 'disappointed' after talks with stricken parent firm Federal Mogul.|\\
    \lstinline|LABEL: Business |\\
    \lstinline|EXPLANATION: The statement discusses talks between unions and a parent firm, which relates to business-related negotiations and concerns regarding pensions. |\\
    \\
    \lstinline|SENTENCE: [{text}]|\\
    \lstinline|EXPLANATION:|
\end{mybox}

\subsection{Baselines with Pairwise Reduction}

Here, we present the pairwise prompt used in baselines with label reduction.
The LLMs are instructed to compare options in pairs within the refined set obtained from the first stage.

\begin{mybox}[Zero-Shot Pairwise Prompt ]
    \lstinline|Which term is more likely to represent the topic of "[{text}]" - "[{label1}]" or "[{label2}]"? |
\end{mybox}

\begin{mybox}[Zero-Shot-CoT Pairwise Prompt ]
    \lstinline|Which term is more likely to represent the topic of  "[{text}]" - "[{label1}]" or "[{label2}]"? |\\
    \lstinline|Let's think step by step and give your explanation to verify your answer: |
\end{mybox}

\begin{mybox}[Few-Shot Pairwise Prompt ]
    \lstinline|Below is a text classification problem, please complete the sentence by "[{label1}]" or "[{label2}]":|\\
    \lstinline|SENTENCE: [{text1}]|\\
    \lstinline|LABEL: [{label1}]|\\
    \\
    \lstinline|... |\\
    \\
    \lstinline|SENTENCE: [{text}]|\\
    \lstinline|LABEL:|
\end{mybox}

\begin{mybox}[Few-Shot-CoT Pairwise Prompt ]
    \lstinline|Below is a text classification problem. Let's think step by step and give your explanation to verify which term is more likely to represent the label of the sentence - "[{label1}]" or "[{label2}]":|\\
    \lstinline|SENTENCE: [{text1}]|\\
    \lstinline|EXPLANATION: [{explain1}]|\\
    \lstinline|LABEL: [{label1}]|\\
    \\
    \lstinline|... |\\
    \\
    \lstinline|SENTENCE: [{text}]|\\
    \lstinline|EXPLANATION:|
\end{mybox}

\subsection{Our PC-CoT Prompt}

Our proposed contrastive elimination methodology, PC-CoT, consists of three key stages: similarity analysis, difference analysis, and final decision-making. 
Specific system instructions and prompts for the three stages are provided below. 

\begin{mybox}[Instruction]
    \lstinline|Below is a text classification problem:|\\
    \lstinline|SENTENCE: [{text1}]|\\
    \lstinline|LABEL: [{label1}]|\\
    \lstinline|SENTENCE:[{text2}]|\\
    \lstinline|LABEL: [{label2}]|
\end{mybox}

\begin{mybox}[Similarity Analysis]
    \lstinline|The phrases can often be mistaken for "[{label1}]" and "[{label2}]", due to certain shared characteristics.|\\
    \lstinline|SHARED ASPECTS: | 
\end{mybox}

\begin{mybox}[Difference Analysis]
    \lstinline|Next, diligently contrast the deviations between these two topics, putting aside the mentioned shared characteristics. Concisely explain, what is the key element that sets them apart? |\\
    \lstinline|CONTRASTING POINTS: | 
\end{mybox}

\begin{mybox}[DecidePrompt]
    \lstinline|After scrutinizing the presented SHARED ASPECTS and CONTRASTING POINTS, which term - "[{label1}]" or "[{label2}]" - would be a more accurate representation for the label of [{text}]? Provide the final label in the format "LABEL: a". |
\end{mybox}

\label{app:os}

% \section{More Open Source Model Result}
% \input{tables/qwen}

% We run experiments on open-source models across different sizes for Qwen models. As shown in Table~\ref{tab:qwen}, We can find that our method still has great advantages for models exceeding 10B, such as Qwen14B. However, for the 1.8B model, our method is lower than FS-CoT, which may be because the smaller model is difficult to induce by our multi-step strategy. Meanwhile, we observe that it also exhibits very bad performance when using FS with full options, we analyze that it stems from the token bias, as illustrated in Figure~\ref{fig:18}. 

\section{Detail of Decision Boundary Ambiguity}

\begin{figure*}[h]
    \centering
    \small
    \includegraphics[width=0.5\linewidth]{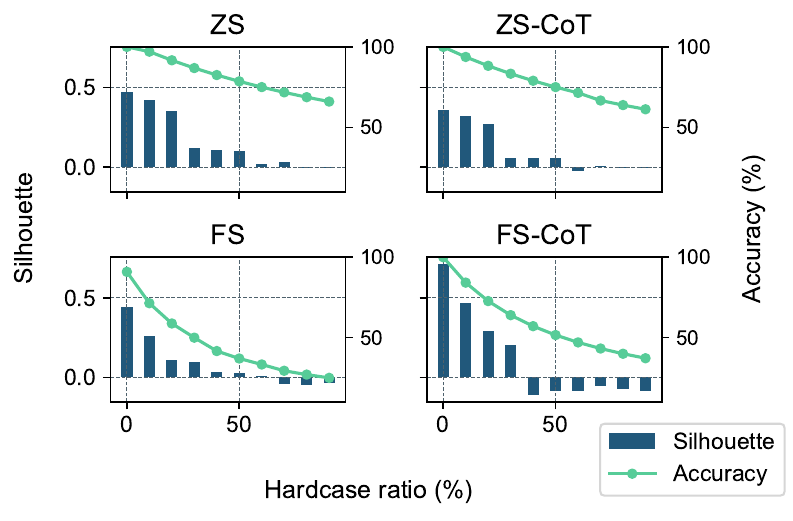}
    \caption{Demonstration sampling methods comparison.}
    \label{fig:sihou}
    % \vspace{-5mm}
\end{figure*}

The silhouette score is a metric used to quantify the compactness and separation of clusters within a dataset. In the context of our study on text data classification, the silhouette score provides insight into the overlap between different class samples. It is calculated for each data point using the formula:

\begin{equation}
s(i) = \frac{b(i) - a(i)}{\max\{a(i), b(i)\}}
\end{equation}

Here, \(a(i)\) is the average distance of a data point \(i\) to other points within the same cluster, and \(b(i)\) is the smallest average distance to points in a different cluster. The overall silhouette score for the dataset is the average of these individual scores, denoted as \(S\):

\begin{equation}
S = \frac{1}{N} \sum_{i=1}^{N} s(i)
\end{equation}
where \(N\) is the total number of data points. So A higher silhouette score indicates clearer clusters and distinct separation suggesting a more evident decision boundary and facilitating classification. Conversely, a lower score suggests increased overlap between class boundaries, indicating a more ambiguous decision boundary.

\begin{figure*}[h]
    \centering
    \begin{subfigure}[h]{0.45\linewidth}
        \centering
        \includegraphics[width=\linewidth]{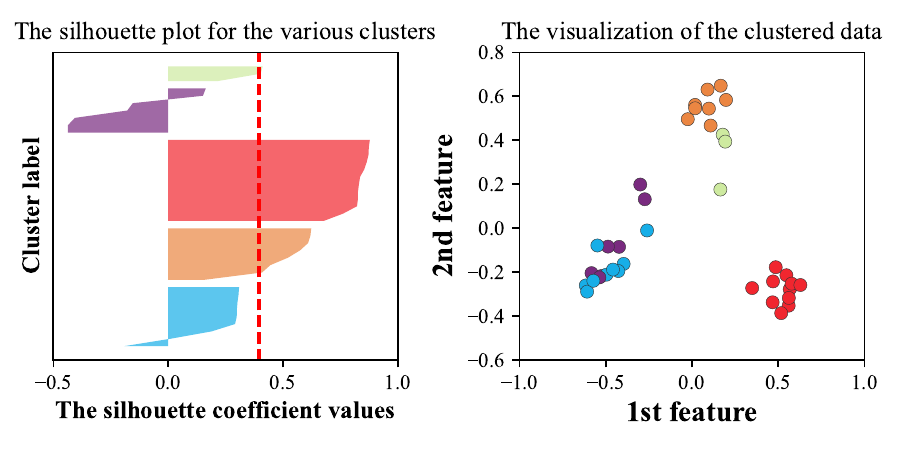}
        \caption{ A higher silhouette score indicates clear boundaries between classes \label{fig:s0}}
    \end{subfigure}
    \begin{subfigure}[h]{0.45\linewidth}
        \centering
        \includegraphics[width=\linewidth]{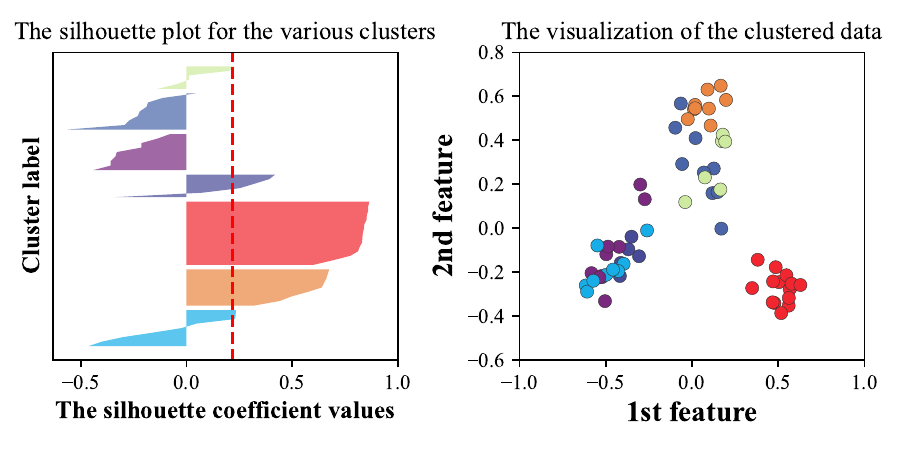}
        \caption{ A lower silhouette score suggesting overlapping classes. \label{fig:s1}}
    \end{subfigure}
    \caption{ Visualization of silhouette scores and decision boundaries }
    \label{fig:two}
\end{figure*}

% In our study, silhouette scores are used to study the relationship between the ambiguous text categories and the decision accuracy of LLM. We observe that the proliferation of options and semantic similarity among them contribute to category overlap and blurred decision boundaries, leading to a rapid degradation in the performance of existing prompting methods.
% This observation motivates us to design a two-stage classification framework, incorporating novel strategies such as self-reduction and our PC-CoT method.

Figure \ref{fig:two} conducts an ablation study on the number of categories. From left to right, we add a new category (in dark blue), leading to increased overlap with existing categories (particularly orange, green, and light blue), resulting in ambiguous boundaries. This overlap is reflected in the decreasing silhouette score, highlighted by the red line. 
Notably, Figure \ref{fig:sihou} demonstrates a clear positive correlation between silhouette score and accuracy, motivating our introduction of a self-reduction stage to decrease the number of categories.
% As only one new categories is introduced and the other clusters are frozened, so the two figure don’t exhibit big differences.

\input{figures/confusion_matrix}

\section{The LIU54 Dataset}
\label{app:liu}

The challenge posed by the LIU54 dataset, as evidenced by the difficulty encountered by even advanced language models like gpt-3.5-turbo and Qwen-72B, underscores a significant obstacle in natural language understanding. Despite the formidable capabilities of these models, they struggle to achieve satisfactory performance, often achieving accuracy rates of only around 30\% using conventional baselines. A closer examination of LIU54 reveals that its difficulty primarily stems from the brevity of its sentences, which hinders effective interpretation and reasoning. This observation underscores the importance of enhancing the capacity of large language models to capture subtle semantic nuances, particularly in the context of classification tasks. Overall, this highlights a critical area for improvement in the development of language understanding technologies.

\begin{mybox}[Semantic incompleteness]
\textbf{sentence}: search \\
\textbf{golden}: \textcolor{red}{factoid} \\
\textbf{prediction}: \textcolor{blue}{query} \\
\\
\textbf{sentence}: Investing \\
\textbf{golden}: \textcolor{red}{quirky} \\
\textbf{prediction}: \textcolor{blue}{stock} 
\end{mybox}
    
\begin{mybox}[Label mismatch]
\textbf{sentence}: What's a parsec? \\
\textbf{golden}: \textcolor{red}{factoid} \\
\textbf{prediction}: \textcolor{blue}{definition} \\
\\
\textbf{sentence}: how we calculate \\
\textbf{golden}: \textcolor{red}{sendemail} \\
\textbf{prediction}: \textcolor{blue}{maths} \\
\\
\textbf{sentence}: Add that song to my playlist. \\
\textbf{golden}: \textcolor{red}{likeness} \\
\textbf{prediction}: \textcolor{blue}{Music} 
\end{mybox}

\begin{mybox}[Multi-label]
\textbf{sentence}: Set a notification for sports game \\
\textbf{golden}: \textcolor{red}{set} \\
\textbf{prediction}: \textcolor{blue}{game} \\
\\
\textbf{sentence}: Is the brightness of my screen running low? \\
\textbf{golden}: \textcolor{red}{quirky} \\
\textbf{prediction}: \textcolor{blue}{hue\_lighton}\\
\\
\textbf{sentence}: What time does the downtown parade start? \\
\textbf{golden}: \textcolor{red}{query} \\
\textbf{prediction}: \textcolor{blue}{events} 
\end{mybox}

We analyzed failed predictions by various Language Model (LLM) architectures using a confusion matrix (see Fig.  ~\ref{fig:liu}) and presented selected failed cases above. Our findings reveal striking similarities in the confusing classes identified by the three LLMs. We categorize the failed cases into three overarching types: (i) Semantic Incompleteness, (ii) Label Mismatch, and (iii) Text Contents of Multiple Classes.

Semantic incompleteness refers to grammatically incomplete sentences typically comprised of a few words, such as 'search', 'Investing', 'Bingo', etc. Label mismatch occurs when the assigned label from the original dataset is inappropriate; for example, 'What's a parsec?' is more fittingly categorized as "definition" rather than "factoid". In cases of multi-label content, sentences express multiple aspects that satisfy multiple categories concurrently. For instance, 'What time does the downtown parade start?' fulfills both the 'query' and 'events' class criteria simultaneously.

Based on our analyses, we posit that confusing labels within the test set of the LIU54 Dataset may introduce bias into results. However, it also presents an opportunity for evaluating the compatibility of different LLMs, thus serving as a challenging benchmark task.

% \section{Simple Dataset}
% \label{app:sim}

% \input{tables/easy.tex}

\section{Case Study}

We present the complete set of options and detailed cases for the pairwise comparison framework, ranging from Table~\ref{tab:case1} to Table~\ref{tab:case9}, along with additional reduction cases outlined in Table~\ref{tab:case10}.

\label{app:case}
\input{tables/pair_examples}

% \section{The Impact of Demonstration Number}

% \input{figures/fewshot.tex}

% We also notice an interesting phenomenon concerning the difference between the Few-Shot and CoT-based methods.
% With an increase in the number of demonstrations, the accuracy of the CoT-based method initially rises, 
% then declines around a shot of 5, and exhibits better performance with a shot number closer to 10.
% In contrast, Few-Shot methods may have a different pattern.
% This may be because the Few-Shot method needs a sufficient number of demonstrations to discern the underlying patterns in text-label relationships. 
% Unlike Few-Shot methods, which rely solely on context, CoT-based methods incorporate their rational process. 
% This may lead to potential fluctuation related to unstable distribution shifts as the number of demonstrations increases. However, it is anticipated that the CoT-based method gradually aligns with the general pattern as the demonstration set becomes more akin to the global distribution.

% In contrast to Few-Shot methods that rely solely on context, CoT-based approaches integrate their reasoning processes. 
% As the number of demonstrations increases, there is a potential for fluctuations associated with unstable distribution changes. 
% For example, randomly sampled demonstrations may conflict with the world knowledge of LLMs, thereby impacting their reasoning processes. 
% Nevertheless, with a sufficient demonstration set, CoT-based methods progressively learn to bridge the distribution gap and converge toward a more generalized pattern.

% TODO: cofusion matrix

\end{document}

%% file: algos/compair.tex
\begin{minipage}{0.95\columnwidth}
\begin{algorithm}[H]\small
\caption{PC-CoT}\label{alg:maccot}
\begin{algorithmic}[1]
\Require Input $x$, reduced label options $\mathcal{R}=\{y_1,y_2,\cdots,y_r\}$, language model $p_\theta$, training set $\mathcal{D}$,
similar analysis prompt \texttt{Sim}, difference analysis prompt \texttt{Diff} and decision prompt \texttt{Dec}.
% run reduced
\While {$|\mathcal{R}| > 1$}
\State Select two label options $\{y_1, y_2\}$ from $\mathcal{R}$
\State $\mathcal{R} = \mathcal{R} \backslash \{y_1, y_2\}$
\State Sample demonstrations pairs $\{x,y\} \in \mathcal{D}$ for $\{y_1, y_2\}$ as $S_{y_1}, S_{y_2}$
\State $S \leftarrow S_{y_1} \cup S_{y_2} $
\State Sample $z_{s} \sim p_\theta^{\texttt{Sim}}(z \mid x, S, \{y_1,y_2\})$
\State Sample $z_{d} \sim p_\theta^{\texttt{Diff}}(z \mid z_{s}, x, S, \{y_1,y_2\})$
\State Predict $y_{p} \sim p_\theta^{\texttt{Dec}}( y \mid z_{s}, z_{d}, x, S , \{y_1, y_2\})$
\State $\mathcal{R} = \mathcal{R} \cup \{y_{p}\}$
\EndWhile
\State \textbf{return} $R$
\end{algorithmic}
\end{algorithm}
\end{minipage}
\hfill

%% file: tables/reducer.tex
\begin{table}[t]
    \centering
    \resizebox{\columnwidth}{!}{
    \begin{tabular}{llcccccccccccccccccccccccccccc}
    \toprule
    \multicolumn{1}{c}{\textbf{Model}} &\textbf{Dataset} & \textbf{P} & \textbf{SC} & \textbf{ITR} & \textbf{CBWR} \\ \midrule
\multirow{5}{*}{\textbf{gpt-3.5-turbo}} 
& \textbf{Banking77}&  40.48          & 65.35         & \textbf{80.05}             & \underline{77.02} \\
& \textbf{HWU64 }   &  49.93          & 76.25         & \underline{90.96}          & \textbf{90.96} \\
& \textbf{LIU54}    &  32.17          & 37.99         & \underline{52.51}          & \textbf{55.69} \\
& \textbf{Clinc150} &  66.11          & 88.28         & \underline{93.11}          & \textbf{94.37} \\
& \textbf{Average}  &  47.17          & 66.97         & \underline{79.16}          & \textbf{79.51} \\ \midrule 
\multirow{5}{*}{\textbf{LLaMA2-70B-Chat}} 
& \textbf{Banking77}&   33.44	    &  37.34       & \textbf{71.43}    & \underline{61.36} \\
& \textbf{HWU64 }   &   39.09	    &  27.27       & \textbf{81.82}    & \underline{78.90}  \\
& \textbf{LIU54}    &   35.04	    &  27.95       & \textbf{63.78}    & \underline{49.61} \\
& \textbf{Clinc150} &   52.44	    &  37.33       & \textbf{85.78}    & \underline{73.99} \\
& \textbf{Average}  &   40.00	    &  32.47       & \textbf{75.70}    & \underline{65.97} \\ \midrule 
    \multirow{5}{*}{\textbf{Qwen-72B-Chat}} 
& \textbf{Banking77}&    42.21     &  71.75     & \textbf{80.52}     & \underline{79.55} \\
& \textbf{HWU64 }   &    56.36     &  72.73     & \underline{85.45}  & \textbf{91.82} \\
& \textbf{LIU54}    &    31.50     &  53.94     & \textbf{63.78}     & \underline{58.27} \\
& \textbf{Clinc150} &    61.78     &  86.67     & \underline{93.33}  & \textbf{94.67} \\
& \textbf{Average}  &    47.96     &  71.27     & \underline{80.77}  & \textbf{81.08} \\ \bottomrule 
    \end{tabular}%
    }
\caption{ HIT@5 performance evaluation for the reduction stage. We compare the effectiveness of our two self-reduction strategies, ITR and CBWR, against LLM employing either a standard reduction prompt (P) or paired with a self-consistency technique (SC). The best results are highlighted in \textbf{bold}, and the second best results are \underline{underlined}. }
\label{tab:red}
\end{table}

%% file: tables/compair.tex
\begin{table*}[t]
\resizebox{\linewidth}{!}{ 
\begin{tabular}{llccccccccccccccccccccccccccc}
\toprule
\multicolumn{1}{c}{\multirow{2}{*}{\textbf{Model}}} & \multicolumn{1}{l}{\multirow{2}{*}{\textbf{Dataset}}} & \multicolumn{4}{c}{\textbf{Full Options}}  &  \multicolumn{5}{c}{\textbf{ Our Framework }}                    \\
\cmidrule(r){3-6} \cmidrule(r){7-11} 
 &  & \textbf{ZS}    & \textbf{ZS-CoT}   & \textbf{FS}     & \textbf{FS-CoT}       & \textbf{ZS} & \textbf{ZS-CoT} & \textbf{FS} & \textbf{FS-CoT}& \textbf{PC-CoT}\\ \midrule
   \multirow{5}{*}{\textbf{gpt-3.5-turbo}} & 
   \textbf{Banking77} & {30.81}\textsubscript{±2.80}  & {37.39}\textsubscript{±1.97}  & {41.91}\textsubscript{±2.97}  & {47.40}\textsubscript{±0.65} & {44.43}\textsubscript{±2.22}  & {45.19}\textsubscript{±3.09}  & \underline{{50.06}}\textsubscript{±4.33}  & {47.53}\textsubscript{±2.87}  & {\textbf{61.82}}\textsubscript{±2.57} \\
 & \textbf{HWU64 }    & {41.48}\textsubscript{±2.88}    & \underline{{60.59}}\textsubscript{±1.99}    & {56.05}\textsubscript{±1.34}    & {57.54}\textsubscript{±4.89}    & {52.43}\textsubscript{±0.11}  & {54.46}\textsubscript{±0.25} & {59.05}\textsubscript{±0.22}  & {58.79}\textsubscript{±0.27}    & {\textbf{65.09}}\textsubscript{±0.87} \\
 & \textbf{LIU54}     & {29.30}\textsubscript{±0.42}  & {31.83}\textsubscript{±1.17}    & 36.37\textsubscript{±0.17}  & {35.39}\textsubscript{±2.76} & {33.95}\textsubscript{±2.62}    & {32.34}\textsubscript{±1.91}  & \underline{{36.39}}\textsubscript{±3.38}  & {31.67}\textsubscript{±2.35}    & {\textbf{37.35}}\textsubscript{±3.75} \\
 & \textbf{Clinc150}  & {59.03}\textsubscript{±1.37}    & {68.61}\textsubscript{±0.55}  & {78.77}\textsubscript{±3.16}    & {76.71}\textsubscript{±0.47} & {77.68}\textsubscript{±3.25}  & {72.12}\textsubscript{±0.17}    & {78.75}\textsubscript{±3.54}    & \underline{{81.29}}\textsubscript{±2.12}  & {\textbf{83.78}}\textsubscript{±3.84} \\
 & \textbf{Average}   & {40.24}\textsubscript{±1.24} & {49.60}\textsubscript{±1.42} & {53.27}\textsubscript{±1.91} & {54.26}\textsubscript{±2.19} & {52.12}\textsubscript{±2.05} & {51.03}\textsubscript{±1.35} & \underline{{56.06}}\textsubscript{±2.87} & {54.82}\textsubscript{±2.33} & {\textbf{62.01}}\textsubscript{±2.33} \\ \midrule
   \multirow{5}{*}{\textbf{LLaMA2-70B-Chat}} & 
   \textbf{Banking77} & {32.04}\textsubscript{±4.55}  & {32.14}\textsubscript{±3.19}  & {27.27}\textsubscript{±3.10}  & {35.71}\textsubscript{±4.50} & {41.56}\textsubscript{±1.26}  & {27.93}\textsubscript{±0.31}  & \textbf{50.54}\textsubscript{±2.28}  & {31.60}\textsubscript{±1.60}  & \underline{49.91}\textsubscript{±1.01} \\
 & \textbf{HWU64 }    & {43.94}\textsubscript{±5.48}  & { 7.57}\textsubscript{±0.53}  & {31.21}\textsubscript{±6.82}  & {42.73}\textsubscript{±2.73} & {50.91}\textsubscript{±1.99}  & {30.31}\textsubscript{±0.52}  &        {61.21}\textsubscript{±2.78}  & {40.61}\textsubscript{±3.20}  & \textbf{50.61}\textsubscript{±1.39} \\
 & \textbf{LIU54}     & {29.13}\textsubscript{±1.80}  & {10.37}\textsubscript{±7.04}  & { 8.66}\textsubscript{±3.61}  & {30.05}\textsubscript{±2.98} & {33.86}\textsubscript{±1.68}  & {25.98}\textsubscript{±0.68}  &        {36.22}\textsubscript{±6.58}  & {29.26}\textsubscript{±0.23}  & \textbf{38.97}\textsubscript{±2.19} \\
 & \textbf{Clinc150}  & {43.70}\textsubscript{±2.72}  & { 5.19}\textsubscript{±1.03}  & {38.67}\textsubscript{±9.69}  & {52.74}\textsubscript{±9.90} & {68.44}\textsubscript{±1.77}  & {42.07}\textsubscript{±0.25}  & \textbf{82.22}\textsubscript{±3.11}  & {55.26}\textsubscript{±2.00}  & \underline{73.33}\textsubscript{±2.10} \\
 & \textbf{Average}   & {37.20}\textsubscript{±3.64}  & {13.82}\textsubscript{±2.95}  & {26.45}\textsubscript{±5.80}  & {40.31}\textsubscript{±5.03} & {48.69}\textsubscript{±1.67}  & {31.57}\textsubscript{±0.44}  & \textbf{57.55}\textsubscript{±3.69}  & {39.18}\textsubscript{±1.76}  & \underline{53.21}\textsubscript{±1.67} \\ \midrule
   \multirow{5}{*}{\textbf{Qwen-72B-Chat}} & 
   \textbf{Banking77} & {43.83}\textsubscript{±1.84}  & {42.05}\textsubscript{±0.23}   & {48.70}\textsubscript{±3.21}  & {47.89}\textsubscript{±2.06}  & {49.68}\textsubscript{±0.46} & {44.97}\textsubscript{±0.23}  &           {53.09}\textsubscript{±4.82} & \underline{60.07}\textsubscript{±5.06}  & \textbf{61.69}\textsubscript{±0.92}\\
 & \textbf{HWU64 }    & {54.55}\textsubscript{±2.57}  & {56.82}\textsubscript{±4.50}   & {55.00}\textsubscript{±0.64}  & {57.73}\textsubscript{±0.64}  & {43.64}\textsubscript{±0.53} & {47.73}\textsubscript{±0.64}  & \underline{60.46}\textsubscript{±3.22} &           {56.82}\textsubscript{±0.64}  & \textbf{63.64}\textsubscript{±1.29} \\
 & \textbf{LIU54}     & {32.68}\textsubscript{±5.56}  & {31.10}\textsubscript{±1.67}   & {40.36}\textsubscript{±4.18}  & {36.62}\textsubscript{±0.56}  & {31.70}\textsubscript{±0.28} & {29.53}\textsubscript{±0.15}  & \underline{39.37}\textsubscript{±2.79} &           {38.78}\textsubscript{±0.28}  & \textbf{41.73}\textsubscript{±1.12} \\
 & \textbf{Clinc150}  & {57.11}\textsubscript{±0.95}  & {58.00}\textsubscript{±2.21}   & {73.56}\textsubscript{±2.20}  & {73.56}\textsubscript{±4.08}  & {67.34}\textsubscript{±0.32} & {63.12}\textsubscript{±0.63}  &           {63.11}\textsubscript{±3.78} & \underline{83.78}\textsubscript{±0.31}  & \textbf{84.89}\textsubscript{±0.49} \\
 & \textbf{Average}   & {47.04}\textsubscript{±2.73}  & {46.99}\textsubscript{±2.15}   & {54.40}\textsubscript{±2.56}  & {53.95}\textsubscript{±1.84}  & {48.09}\textsubscript{±0.40} & {46.34}\textsubscript{±0.41}  &           {54.01}\textsubscript{±3.65} & \underline{59.86}\textsubscript{±1.57}  & \textbf{62.99}\textsubscript{±0.95} \\ \bottomrule
\end{tabular}%
}
\caption{ Accuracy performance evaluation of three LLMs on four challenge benchmarks. We experiment with various LLM prompting techniques: Zero-Shot (ZS), Few-Shot (FS), Zero-Shot-CoT (ZS-CoT), and Few-Shot-CoT (FS-CoT).
We apply these techniques either with full options or within our reduction-then-pairwise-comparison framework.
Our comparison algorithm, PC-CoT, is used within our framework.
For gpt-3.5-turbo, we employ 3-shot for fewshot demonstrations, while for LLaMA2 and Qwen, we use 5-shot.  
The best results are highlighted in \textbf{bold}, and the second best results are \underline{underlined}.  }
\label{tab:com}
\end{table*}

%% file: tables/ablation.tex
\begin{table}[ht]
\small
\centering
\resizebox{0.9\columnwidth}{!}{ 
\begin{tabular}{lcccccccccccccccccccccccccccc}
\toprule
{\textbf{Dataset}} &\textbf{PC-CoT} & \textbf{w/o S}& \textbf{w/o D}& \textbf{w/o SD} \\ \midrule
\textbf{Banking77} & {\textbf{61.82}}& {51.53} & {50.18} & {47.34} \\
\textbf{HWU64 }   & {\textbf{65.09}} & {52.44}  & {61.11}  & {59.01}  \\
\textbf{LIU54}    & {\textbf{37.35}} & {30.31}  & {31.34}  & {33.37}  \\
\textbf{Clinc150} & {\textbf{83.78}} & {75.68}  & {80.22}  & {74.65}  \\
\textbf{Average}  & {\textbf{62.01}} & {52.49} & {55.71} & {52.59} \\ \bottomrule
\end{tabular}%
}
\caption{The ablation study for our PC-CoT algorithm (with gpt-3.5-turbo). We evaluate the performance under various conditions: remove the similar analysis prompt (w/o S), remove the difference analysis prompt (w/o D), and remove both similar and difference analysis prompt (w/o SD). }
\label{tab:ab}
\end{table}

%% file: tables/sft.tex
\begin{table*}[h]
    \centering
    \resizebox{\linewidth}{!}{%
    \begin{tabular}{@{}llccccccccccccccccccccccccccc@{}}
    \toprule
{{\textbf{Shot}}} &  {\textbf{Dataset}} & {\textbf{BERT-base}} & {\textbf{RoBerta-base}} & {\textbf{BERT-Large}} & {\textbf{RoBerta-large}} & {\textbf{Llama-1.1B}}  & {\textbf{Qwen-1.8B}} \\ \midrule 
\multirow{5}{*}{\textbf{K=1}}
& \textbf{Banking77}  & 1.95   & 1.30    & 2.60    & 4.22  & 28.90    & 53.25     \\
& \textbf{Clinc150}   & 1.33   & 0.89    & 1.78    & 6.67  & 45.78    & 60.44     \\
& \textbf{LIU54}      & 2.76   & 4.33    & 7.09    & 4.33  & 24.02    & 39.37     \\
& \textbf{HWU64}      & 5.45   & 1.82    & 0.91    & 3.64  & 16.36    & 53.64     \\
& \textbf{Avg.}       & 2.87   & 2.08    & 3.09    & 4.71  & 28.76    & 51.67     \\ \midrule
\multirow{5}{*}{\textbf{K=5}}
& \textbf{Banking77}  & 0.32                & 6.49              & 1.30            & 0.32            & 40.58              & 70.45                \\
& \textbf{Clinc150}   & 7.56                & 36.44             & 15.11            & 47.56            & 68.00              & 79.56                \\
& \textbf{LIU54}      & 8.66                & 16.14             & 9.84            & 14.17            & 29.92              & 50.39                \\
& \textbf{HWU64}      & 6.36                & 18.18             & 21.82            & 32.73            & 47.27              & 65.45                \\
& \textbf{Avg.}       & 5.73                & 19.32             & 12.02            & 23.70            & 46.44              & 66.46                \\ \bottomrule
\end{tabular}%
    }
    \caption{ Accuracy performance evaluation of encoder-only and decoder-only models. }
    \label{tab:sft}
    \vspace{-3mm}
    \end{table*}

%% file: algos/reducer2.tex
% \begin{minipage}{0.91\columnwidth}
%     \begin{algorithm}[H]\small
%     \caption{Self-Reduction (CBWR)}\label{alg:slidewindow}
%     \begin{algorithmic}[1]
%     \Require Input sentence $x$, label options set $\mathcal{Y}= \{y_1,y_2,\cdots,y_N\}$, LLM $p_\theta$, evaluator $E()$, selection prompt \texttt{Sel}, step limit T.
%     \State Initialize $\mathcal{S}=\{\}$, $\mathcal{E}=\{\}$
%     \For {each options $y \in \mathcal{S}$}
%     \State $e \leftarrow \text{SBERT}(y) $
%     \State $\mathcal{E} \leftarrow \mathcal{E} \cup \{e\}$
%     \EndFor
%     \For {$t = 1,\cdots,T$}
%     % cluster
%     \State $\mathcal{E} \leftarrow \{e_i | i \in \mathcal{S}\}$
%     \State Cluster $\mathcal{E}$ into $K$ cluster.
%     % windows 
%     \For {each cluster $i = 1, \cdots, K$}
%     \State Initialize $W_i = \{\}$
%     \State Allocate each option in cluster $i$ into the window $W_i$ by Round-Robin algorithm.
%     \EndFor
%     % reduction
%     \State Initialize $\mathcal{R}=\{\}$
%     \For {$i = 1, \cdots, K$}
%     \State $y_w \leftarrow \text{argmax}_y p_\theta^\texttt{Sel}(y | x,W_i)$
%     \State $\mathcal{R} \leftarrow  \mathcal{R}\cup \{y_w\}$
%     \EndFor
%     \State $S_{t+1} \leftarrow  \mathcal{R}$
%     \EndFor
%     \State \textbf{return} $\mathcal{R}$
%     \end{algorithmic}
%     \end{algorithm}
%     \end{minipage}
%     \hfill
%     \vspace{1mm}

\begin{minipage}{0.91\columnwidth}
    \begin{algorithm}[H]\small
    \caption{Self-Reduction (CBWR)}\label{alg:cbwr}
    \begin{algorithmic}[1]
    \Require Input sentence $x$, full options set $\mathcal{Y}= \{y_1,y_2,\cdots,y_N\}$, LLM $p_\theta$, selection prompt \texttt{Sel}, selection number $N$, step limit $T$
    \State Initialize the selection set $\mathcal{S}=\mathcal{Y}$
    \For {$t = 1$ to $T$}
        \State Cluster the option set $\mathcal{S}$ into $K$ clusters
        \State $W \rightarrow \{\}$
        \For {each cluster $i$ from 1 to $K$}
            \State Randomly select options from cluster $i$ 
            \State add them into the window $W$
        \EndFor
        \State Select candidate set $C$ from $W$: 
        \Statex \hspace{\algorithmicindent} $C \leftarrow \text{Top-N}\Big(\{y \mid p_\theta^{\texttt{Sel}}(y \mid x, W)\}\Big)$
        \State Determine the discard option set $\mathcal{D} = W \setminus C$
        \State Update the remaining candidates $\mathcal{S} \leftarrow \mathcal{S} \setminus \mathcal{D}$
    \EndFor
    \State \textbf{return} the remaining selection set $\mathcal{S}$
    \end{algorithmic}
    \end{algorithm}
\end{minipage}
\hfill
\vspace{1mm}

%% file: figures/confusion_matrix.tex
\begin{figure*}[htbp]
    \centering
    % \begin{subfigure}[h]{0.49\textwidth}
    %     \centering
    %     \includegraphics[width=\textwidth]{figures/cm-Bert.pdf}
    % \end{subfigure}
    \begin{subfigure}[h]{0.32\textwidth}
        \centering
        \includegraphics[width=\textwidth]{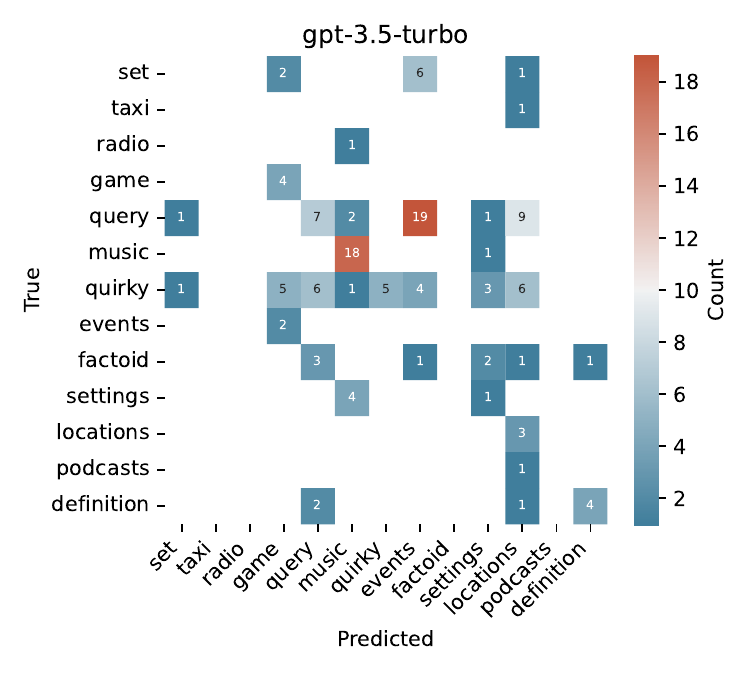}
    \end{subfigure}
    \begin{subfigure}[h]{0.32\textwidth}
        \centering
        \includegraphics[width=\textwidth]{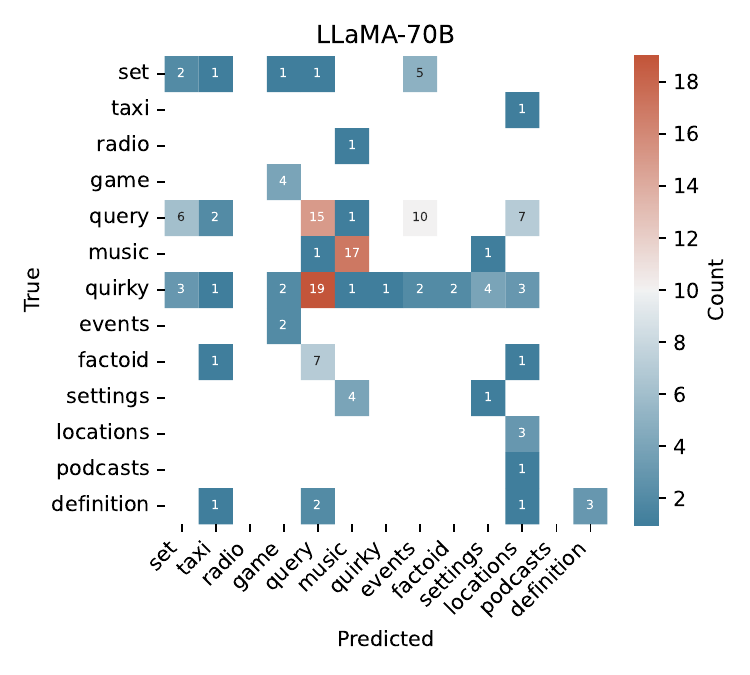}
    \end{subfigure}
    \begin{subfigure}[h]{0.32\textwidth}
        \centering
        \includegraphics[width=\textwidth]{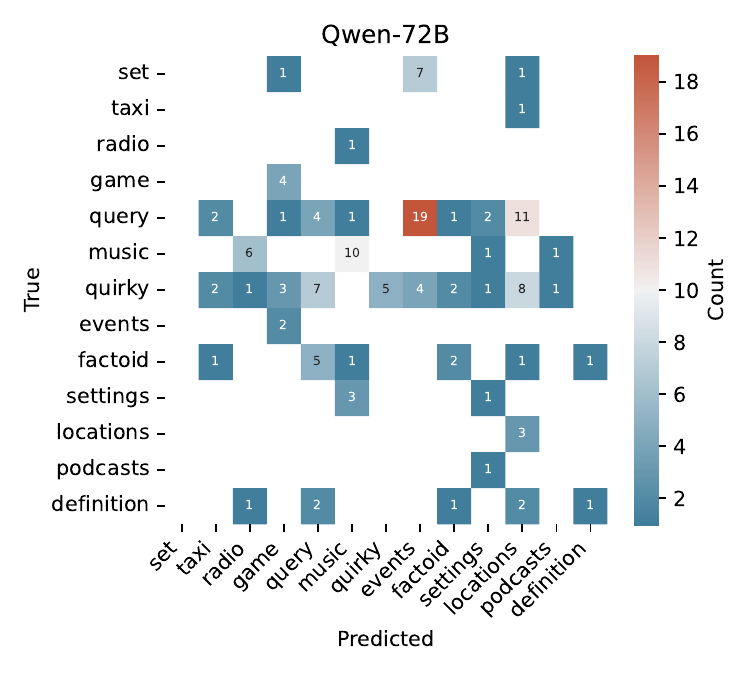}
    \end{subfigure}
    
    \caption{Visualization of bad cases of three LLMs on LIU54.}
    \label{fig:liu}
\end{figure*}

%% file: tables/pair_examples.tex
\begin{table*}[h]\centering
    \resizebox{\linewidth}{!}{%
    \begin{tabular}[t]{p{1\textwidth}}
    \toprule 
    \textbf{ZERO-SHOT } \\
    \midrule
    \bblack{Q:} Given the sentence: "So I just put my top-up into the card and it hasn't changed." Please select the most possible topic from the following options.\\
    \bblack{OPTIONS:} pending\_top\_up, top\_up\_failed, topping\_up\_by\_card, card\_not\_working, balance\_not\_updated, ...\\
    \bblack{CHOICE:} \\
    \\
    \bblack{A:} The most possible topic for the given sentence is \textbf{topping\_up\_by\_card}. \ngmark \\
    \bottomrule
    \end{tabular}}
    \caption{Zero-shot reasoning example for the classification problem on the BANKING77 dataset using the gpt-3.5-turbo model.}
    \label{tab:case1}
\end{table*}

\begin{table*}[h]\centering
    \resizebox{\linewidth}{!}{%
    \begin{tabular}[t]{p{1\textwidth}}
    \toprule
    \textbf{FEW-SHOT } \\
    \midrule
    \bblack{Q:} Below is a text classification problem, Note that you can only select labels in pending\_top\_up, top\_up\_failed, topping\_up\_by\_card, card\_not\_working, balance\_not\_updated, ... \\
    \bblack{SENTENCE:} How long will it take for my cash withdrawal to no longer show as pending? \\
    \bblack{LABEL:} pending\_cash\_withdrawal \\
    ... \\
    \bblack{SENTENCE:} So I just put my top-up into the card and it hasn't changed. \\
    \bblack{LABEL:} \\
    \\
    \bblack{A:} \textbf{card\_not\_working} \ngmark\\
    \bottomrule
    \end{tabular}}
    \label{tab:case}
    \caption{Few-shot reasoning example for the classification problem on the BANKING77 dataset using the gpt-3.5-turbo model.}
\end{table*}

\begin{table*}[h]\centering
    \resizebox{\linewidth}{!}{%
    \begin{tabular}[t]{p{1\textwidth}}
    \toprule
    \textbf{ZERO-SHOT COT } \\
    \midrule
    \bblack{Q:} Given the sentence: "So I just put my top-up into the card and it hasn't changed." Please select the most possible topic from the following options.\\
    \bblack{OPTIONS:} pending\_top\_up, top\_up\_failed, topping\_up\_by\_card, card\_not\_working, balance\_not\_updated, ...\\
    Let's think step by step and verify your answer: \\
    \\
    \bblack{A:}Based on the given sentence "So I just put my top-up into the card and it hasn't changed," the most relevant topic seems to be \textbf{top\_up\_failed.} This is because the user mentions putting a top-up into the card, and there is an indication that something has not changed as expected. \ngmark\\
    \bottomrule
    \end{tabular}}
    \label{tab:case}
    \caption{Zero CoT reasoning example for the classification problem on the BANKING77 dataset using the gpt-3.5-turbo model.}
\end{table*}

\begin{table*}[h]\centering
    \resizebox{\linewidth}{!}{%
    \begin{tabular}[t]{p{1\textwidth}}
    \toprule
    \textbf{FEW-SHOT COT } \\
    \midrule
    \bblack{Q:} Below is a text classification problem, Note that you can only select labels in pending\_top\_up, top\_up\_failed, topping\_up\_by\_card, card\_not\_working, balance\_not\_updated, ... Let's think step by step and give your explanation to verify the answer. \\
    \bblack{SENTENCE:} I topped up but it isn't in my account \\
    \bblack{EXPLANATION:} The statement is about a user's concern regarding a top-up that has not been reflected in their account, which is related to a pending transaction or issue with their account balance. \\
    \bblack{LABEL:} pending\_top\_up \\
    ... \\
    \bblack{SENTENCE:} So I just put my top-up into the card and it hasn't changed. \\
    \bblack{EXPLANATION:} \\
    \\
    \bblack{A:} The sentence is about a user who has topped up their card with money, but the balance has not been updated. This issue is related to the balance not being updated after a bank transfer, which falls under the label balance\_not\_updated\_after\_bank\_transfer. \\
    LABEL: \textbf{balance\_not\_updated} \ngmark \\
    \bottomrule
    \end{tabular}}
    \label{tab:case}
    \caption{Few-shot CoT reasoning example for the classification problem on the BANKING77 dataset using the gpt-3.5-turbo model.}
\end{table*}

\begin{table*}[h]\centering
    \resizebox{\linewidth}{!}{%
    \begin{tabular}[t]{p{1\textwidth}}
    \toprule 
    \textbf{ZERO-SHOT PAIRWISE } \\
    \midrule
    \bblack{Q:} Which term is more likely to represent the topic of "So I just put my top-up into the card and it hasn't changed." - "top\_up\_failed" or "pending\_top\_up"?\\
    \\
    \bblack{A:} The term " \textbf{top\_up\_failed}" is more likely to represent the topic of "So I just put my top-up into the card and it hasn't changed." \ngmark \\
    \bottomrule
    \end{tabular}}
    \label{tab:case}
    \caption{Zero-shot pairwise reasoning example for the classification problem on the BANKING77 dataset using the gpt-3.5-turbo model.}
\end{table*}

\begin{table*}[h]\centering
    \resizebox{\linewidth}{!}{%
    \begin{tabular}[t]{p{1\textwidth}}
    \toprule
    \textbf{FEW-SHOT PAIRWISE } \\
    \midrule
    \bblack{Q:} Below is a text classification problem, please complete the sentence by "top\_up\_failed" or "pending\_top\_up": \\
    SENTENCE: I topped up but it isn't in my account \\
    LABEL: pending\_top\_up \\
    ...
     \\
    SENTENCE: So I just put my top-up into the card and it hasn't changed. \\
    LABEL: " \\
    \\
    \bblack{A:} \textbf{top\_up\_failed} \ngmark \\
    \bottomrule
    \end{tabular}}
    \label{tab:case}
    \caption{Few-shot pairwise reasoning example for the classification problem on the BANKING77 dataset using the gpt-3.5-turbo model.}
\end{table*}

\begin{table*}[h]\centering
    \resizebox{\linewidth}{!}{%
    \begin{tabular}[t]{p{1\textwidth}}
    \toprule
    \textbf{ZERO-SHOT COT PAIRWISE } \\
    \midrule
    \bblack{Q:} Which term is more likely to represent the topic of "So I just put my top-up into the card and it hasn't changed." - "top\_up\_failed" or "pending\_top\_up"? Let's think step by step. \\
    \\
    \bblack{A:}Based on the given statement, the term "\textbf{top\_up\_failed}" is more likely to represent the topic. This is because the statement suggests that the top-up process did not work as expected, indicating a potential failure in the top-up transaction. \ngmark \\
    \bottomrule
    \end{tabular}}
    \label{tab:case}
    \caption{Zero-shot CoT pairwise reasoning example for the classification problem on the BANKING77 dataset using the gpt-3.5-turbo model.}
\end{table*}

\begin{table*}[h]\centering
    \resizebox{\linewidth}{!}{%
    \begin{tabular}[t]{p{1\textwidth}}
    \toprule
    \textbf{ZERO-SHOT COT PAIRWISE } \\
    \midrule
    \bblack{Q:} Below is a text classification problem. Let's think step by step and give your explanation to verify which term is more likely to represent the label of the sentence - "top\_up\_failed" or "pending\_top\_up": \\
    SENTENCE: Why has my top up failed? \\
    EXPLANATION: The statement directly mentions a failed top-up, indicating that there was an issue with adding additional funds or credits to something. This is a specific problem related to a top-up process, hence the label "top\_up\_failed." \\
    LABEL: top\_up\_failed \\
    ... \\
    SENTENCE: So I just put my top-up into the card and it hasn't changed. \\
    EXPLANATION: \\
    \\
    \bblack{A:} The statement mentions putting a top-up into a card, but it also mentions that it hasn't changed. This suggests that there may be an issue with the top-up process, indicating a "top\_up\_failed" label. \\LABEL: \textbf{top\_up\_failed} \ngmark \\
    \bottomrule
    \end{tabular}}
    \label{tab:case}
    \caption{Zero-shot CoT pairwise reasoning example for the classification problem on the BANKING77 dataset using the gpt-3.5-turbo model.}
\end{table*}

\begin{table*}[h]\centering
    \resizebox{\linewidth}{!}{%
    \begin{tabular}[t]{p{1\textwidth}}
    \toprule
    \textbf{PC-COT (Ours) } \\
    \midrule
    \bblack{Q:} Below is a text classification problem: \\
    SENTENCE: My top-up didn't go through \\
    LABEL: top\_up\_failed \\
    SENTENCE:why isn't my top-up going through? \\
    LABEL: pending\_top\_up \\
    ... \\
    The phrases can often be mistaken for "top\_up\_failed" and "pending\_top\_up", due to certain shared characteristics.\\
    SHARED ASPECTS:  \\
    \\
    \bblack{A:} - Mention of "top-up" or "top up" \\
    - Mention of the action not going through or failing \\
    - Expressions of confusion or concern about the status of the top-up \\
    \\
    \bblack{Q:}Next, diligently contrast the deviations between these two topics, putting aside the mentioned shared characteristics. Concisely explain, what is the key element that sets them apart?\\
    CONTRASTING POINTS: \\
    \\
    \bblack{A:} The key element that sets apart the "top\_up\_failed" and "pending\_top\_up" topics is the specific focus of the phrases.  \\
    - "top\_up\_failed" phrases primarily focus on the fact that \underline{the top-up has failed or is not working}. They express frustration or confusion about why the top-up did not go through and may ask for reasons or explanations for the failure. \\
    - On the other hand, "pending\_top\_up" phrases primarily focus \underline{on the status or confirmation of the top-up}. They express uncertainty or concern about whether the top-up has gone through or not. These phrases often ask for information about why the top-up is still pending or how to determine if it has been successful. \\
    In summary, the key element that sets them apart is the specific emphasis on the failure of the top-up in "top\_up\_failed" phrases and the uncertainty about the status of the top-up in "pending\_top\_up" phrases.\\
    \\
    \bblack{Q:}After scrutinizing the presented SHARED ASPECTS and CONTRASTING POINTS, which term - "top\_up\_failed" or "pending\_top\_up" - would be a more accurate representation for the label of So I just put my top-up into the card and it hasn't changed.? Provide the final label in the format "LABEL: a".  \\
    \\
    \bblack{A:} 
    Based on the shared aspects and contrasting points provided, the sentence "So I just put my top-up into the card and it hasn’t changed" seems to align more closely with the "pending\_top\_up" label. This is because the sentence \underline{expresses uncertainty or concern} about the status of the top-up, indicating that it hasn't changed and implying that the user is unsure whether the top-up has been successful or not. Therefore, "pending\_top\_up" would be a more accurate representation of the label in this case. LABEL: \textbf{pending\_top\_up} \okmark\\
    \bottomrule
    \end{tabular}}
    \caption{PC-CoT reasoning example for the classification problem on the BANKING77 dataset using the gpt-3.5-turbo model.}
    \label{tab:case9}
\end{table*}

\begin{table*}[t]
\begin{tabular*}{\textwidth}{@{\extracolsep{\fill}}lll}
\midrule
\textbf{Sentence} & \multicolumn{2}{l}{So I just put my top-up into the card and it hasn't changed.}  \\ 
\textbf{Label} & \multicolumn{2}{l}{pending\_top\_up}   \\ \midrule
\textbf{Reduce} & \multicolumn{1}{p{6cm}}{\raggedright \uline{CBWR}} & \multicolumn{1}{p{6cm}}{\raggedright \uline{ITR}} \\
 & \begin{tabular}[c]{@{}l@{}}verify\_top\_up\\      top\_up\_failed\\      balance\_not\_updated\_after\_bank\_transfer\\      pending\_top\_up\\      top\_up\_by\_card\_charge\end{tabular} & \begin{tabular}[c]{@{}l@{}}top\_up\_reverted\\      top\_up\_by\_card\_charge\\      top\_up\_failed\\      balance\_not\_updated\_after\_bank\_transfer\\      topping\_up\_by\_card\end{tabular} \\ \midrule
\textbf{Sentence} & \multicolumn{2}{l}{I made a transfer and am still waiting.}  \\ 
\textbf{Label} & \multicolumn{2}{l}{balance\_not\_updated\_after\_bank\_transfer}   \\ 
\midrule
\textbf{Reduce} & \multicolumn{1}{p{6cm}}{\raggedright \uline{CBWR}} & \multicolumn{1}{p{6cm}}{\raggedright \uline{ITR}} \\
& \begin{tabular}[c]{@{}l@{}}failed\_transfer\\ transfer\_into\_account\\ balance\_not\_updated\_after\_bank\_transfer\\ pending\_transfer\\ transfer\_not\_received\_by\_recipient\end{tabular} & \begin{tabular}[c]{@{}l@{}}pending\_transfer\\ transfer\_not\_received\_by\_recipient\\ transfer\_timing\\ transfer\_fee\_charged\\ failed\_transfer\end{tabular} \\
\midrule
\textbf{Sentence} & \multicolumn{2}{l}{Where can I get my card at?}  \\ 
\textbf{Label} & \multicolumn{2}{l}{order\_physical\_card}   \\ 
\midrule
\textbf{Reduce} & \multicolumn{1}{p{6cm}}{\raggedright \uline{CBWR}} & \multicolumn{1}{p{6cm}}{\raggedright \uline{ITR}} \\
& \begin{tabular}[c]{@{}l@{}}card\_delivery\_estimate\\      card\_arrival\\      getting\_spare\_card\\      order\_physical\_card\\      get\_physical\_card\end{tabular} & \begin{tabular}[c]{@{}l@{}}get\_physical\_card\\      card\_arrival\\      activate\_my\_card\\      card\_delivery\_estimate\\      getting\_virtual\_card\end{tabular} \\
\midrule
\end{tabular*}
\caption{ Examples of CBWR and ITR reduction techniques. The reduction process aims to retain a subset of options with high similarity from a large pool of options as much as possible.}
\label{tab:case10}
\end{table*}